\documentclass[10pt,twocolumn,letterpaper]{article}

\usepackage{cvpr}      
\definecolor{cvprblue}{rgb}{0.21,0.49,0.74}
\usepackage[pagebackref,breaklinks,colorlinks,allcolors=cvprblue]{hyperref}
\usepackage{algorithm}
\usepackage{algorithmic}
\usepackage{booktabs}
\usepackage{multirow}
\usepackage{makecell}   
\usepackage{colortbl}   
\usepackage[table]{xcolor}
\usepackage{array}

\usepackage{threeparttable}
\usepackage{tabularx}

\usepackage{graphicx}
\usepackage{multirow}
\usepackage{amssymb}
\usepackage{pifont}
\definecolor{checkblue}{RGB}{30,100,190}
\definecolor{crossred}{RGB}{190,60,60}
\usepackage{threeparttable}
\usepackage{amsmath} 
\usepackage[nameinlink,capitalize]{cleveref}
\usepackage{newfloat}
\usepackage{listings}
\def\confName{CVPR}
\def\confYear{2027}

\title{SlopeGuard: Towards Generalizable AI-Generated Image Detection via Radial Spectral Slope Guidance}

\author{%
  Cheng Fu\textsuperscript{1,\ensuremath{*}},  
  Yiheng Li\textsuperscript{2,3,\ensuremath{*}},
  Yang Yang\textsuperscript{2,3},
  Wenzheng Liu\textsuperscript{1,\ensuremath{\dagger}}\\
    \textsuperscript{1}School of Computer Science and Technology,  Changsha University of Science and Technology\\
  \textsuperscript{2}School of Artificial Intelligence, University of Chinese Academy of Sciences\\
  \textsuperscript{3}MAIS, Institute of Automation, Chinese Academy of Sciences\\
{\small\texttt{\{202308170227, liuwenzheng\}@csust.edu.cn,
\{liyiheng2024, yangyang2013\}@ia.ac.cn}}%
}

\begin{document}
\maketitle
\begin{abstract}
AI-generated image detectors often overfit to generator-specific artifacts, hindering generalization to unseen synthesis pipelines. Existing frequency-based methods usually learn high-dimensional spectral patterns that may preserve generator-dependent variations. We explore Radial Spectral Slope, a compact descriptor of global spectral decay that captures transferable spectral trends while suppressing fine-grained variations. Our analysis shows that real and generated images exhibit systematic but generator-dependent slope differences, indicating that slope provides useful forensic evidence but is insufficient as a standalone decision criterion. This motivates combining spectral slope with semantic representations for more robust detection. Based on this insight, we propose \textbf{SlopeGuard}, which couples slope-aware data processing with slope guided feature learning to construct informative patch-based views, integrate Radial Spectral Slope with semantic features, and adapt prior preservation according to spectral atypicality. Extensive experiments demonstrate that SlopeGuard achieves state-of-the-art mean accuracies of 97.5\% on AIGCDetectBench, 97.8\% on GenImage, and 85.1\% on AIGIBench, while maintaining competitive performance under blur, cropping, JPEG compression, and additive noise.
\end{abstract}    
\section{Introduction}
\label{sec:intro}
The rapid evolution of image synthesis has turned photorealistic content generation from a specialized capability into a broadly accessible technology~\cite{xu2024ufogen,jia2025chatgen,cavia2024real}. Generative adversarial networks, diffusion models, and personalized generation systems can now produce images that are increasingly difficult to distinguish by visual appearance alone~\cite{goodfellow2014generative,dhariwal2021diffusion,rombach2022high,wu2024infinite}. This progress poses an evolving challenge for AI-generated image detection, requiring detectors to recognize synthetic content from both generators seen during training and previously unseen architectures, sampling procedures, and post-processing pipelines encountered after deployment.~\cite{corvi2023intriguing,li2026artificial,he2026diversity}. We, therefore, focus on forensic evidence that can support detection without requiring the model to preserve every artifact associated with the training generators.

\begin{figure}[t]
    \centering
    \includegraphics[width=\columnwidth]{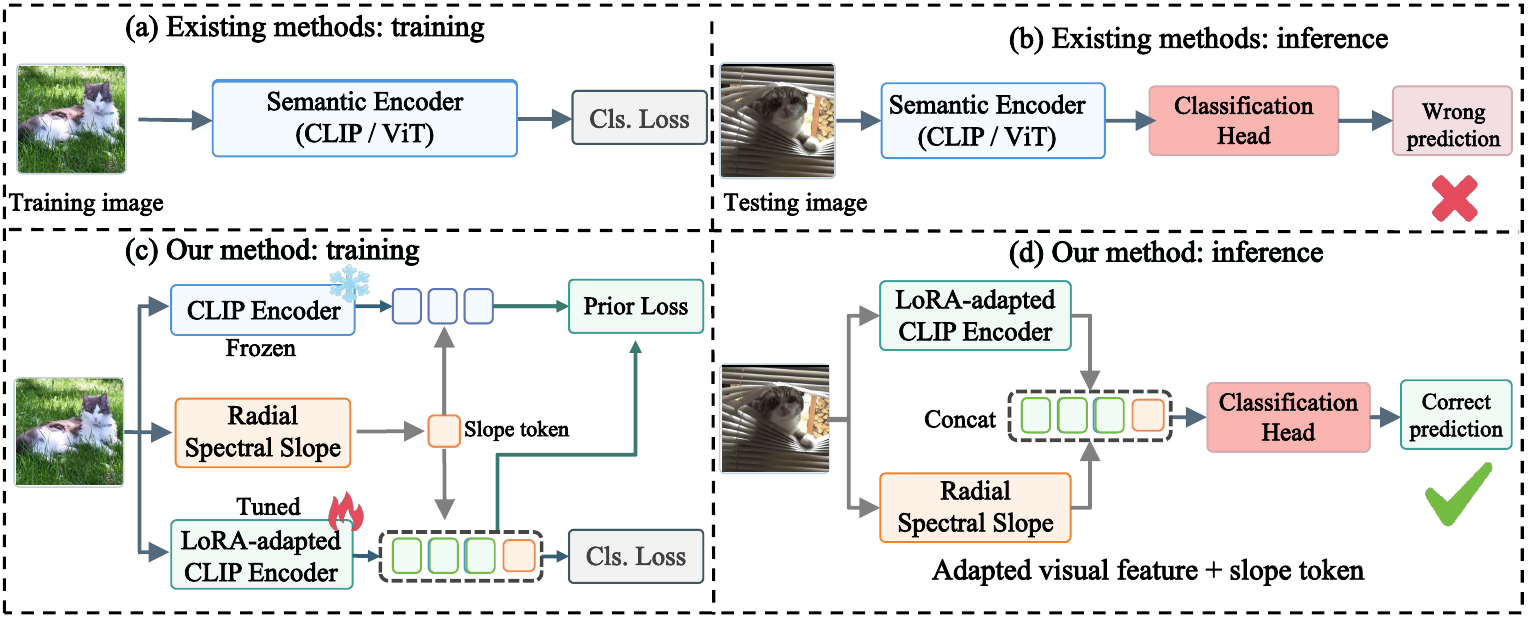}
    \caption{
    Comparison of training and inference pipelines.
    (a,b) Semantic-based detectors optimize visual representations
    through classification supervision and use the learned features
    for authenticity prediction.
    (c,d) SlopeGuard complements LoRA-adapted CLIP features with
    radial spectral slope and uses spectral atypicality to adjust
    prior regularization against a frozen CLIP encoder during training.
    At inference, the frozen branch is removed, while the adapted
    visual features and slope descriptor are concatenated and
    passed to the classification head for authenticity prediction.
    }
    \label{fig:intr}
\end{figure}

Although these approaches have substantially improved cross-generator detection, recent benchmarks reveal notable performance variations across synthesis paradigms, datasets, and image transformations~\cite{yan2025sanity,li2026artificial}.
This suggests that improving generalization may require not only stronger representations but also a reconsideration of the abstraction level at which forensic evidence is captured.
As illustrated in Fig.~\ref{fig:intr}, SlopeGuard complements semantic representations with radial spectral slope $\alpha$, introducing an additional frequency-aware forensic cue for more robust detection across diverse generators.

\begin{figure*}[h]
    \centering
    \includegraphics[width=\linewidth]{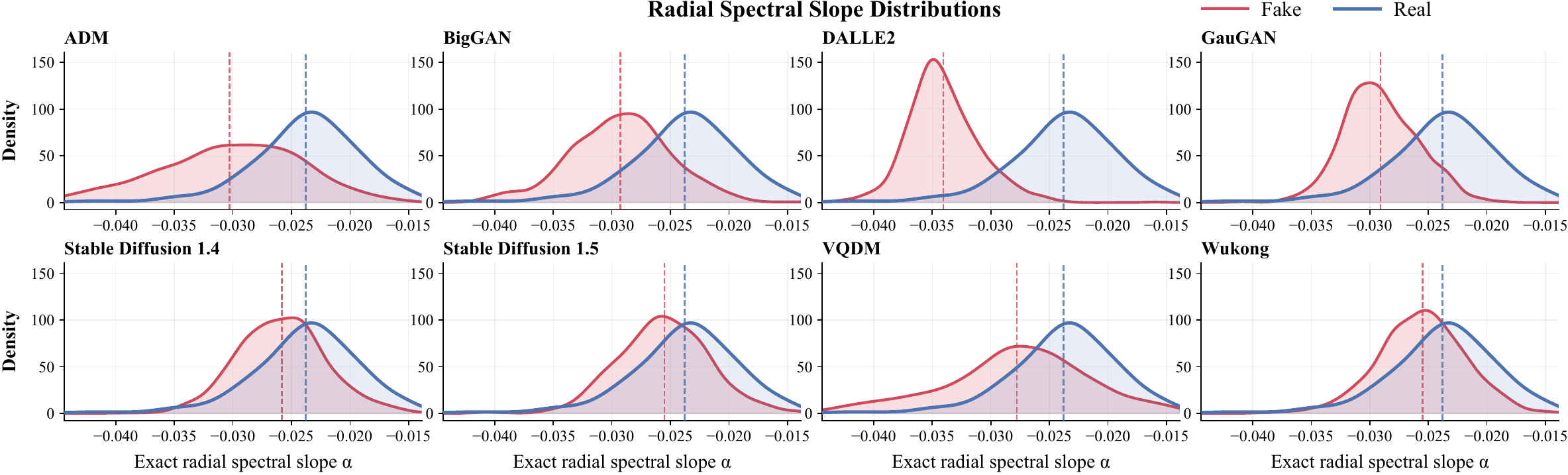}
\caption{Distributions of the exact radial spectral slope $\alpha$ for real and AI-generated images across eight representative generators, using 1,000 real and 1,000 generated images per generator. The distributional differences motivate investigating spectral slopes as auxiliary features of radial spectral slopes for AI-generated image detection.}
    \label{fig:slope_distribution}
\end{figure*}

Frequency-domain forensics makes this representational choice especially visible~\cite{karageorgiou2025any,tan2024frequency}. Individual Fourier responses and wavelet coefficients expose synthesis traces that can be difficult to observe directly in the spatial domain~\cite{frank2020leveraging,durall2020watch,corvi2023detection,wolter2022wavelet}. Dense or learned spectral representations retain these responses for discrimination~\cite{tan2024frequency,yan2026dual,li2025improving} but also preserve detailed variations associated with generator architectures, resampling operators, and image-processing pipelines. Semantic representations operate at a different level of abstraction, providing transferable visual knowledge without explicitly characterizing how spectral energy is distributed across frequencies~\cite{zheng2024breaking,zhou2025aigi}. Recent methods partially bridge this gap by learning the spectral distribution of real images or modeling the elevated frequency spectral tail~\cite{karageorgiou2025any,li2026spectral}. We investigate a complementary direction by examining whether the global spectral decay trend can serve as a compact image-level forensic signal without preserving individual frequency responses.

Generalizable detection requires frequency representations that preserve transferable synthesis-related trends while reducing dependence on generator-specific spectral patterns. We adopt radial spectral slope as this intermediate abstraction. By summarizing the radial decay of log amplitude with a single interpretable coefficient, it suppresses frequency bin variations while retaining the dominant spectral trend. As shown in Fig.~\ref{fig:slope_distribution}, real and generated images exhibit systematic, generator dependent shifts in their slope distributions, with varying degrees of overlap. Thus, the absolute slope provides discriminative evidence but cannot support a reliable universal threshold, motivating its integration with learned visual representations as a complementary cue.

Simply appending the slope to a classifier, however, does not control how the visual representation evolves during adaptation. Training on a limited set of generators may steer the adapted CLIP encoder toward generator specific artifacts and weaken the transferable prior inherited from pretraining. We therefore retain a frozen CLIP branch as a stable reference and regularize the adapted representation toward this prior. A uniform constraint assigns the same preservation strength to all samples without accounting for differences in their radial decay patterns. Motivated by this spectral variation, we investigate batch-relative slope deviation as a signal for modulating the prior constraint. We use this deviation as an empirical proxy for allocating regularization strength, with controlled weighting comparisons providing evidence for the effectiveness of this design. Samples with more atypical slopes receive stronger but bounded preservation, while normalization maintains a stable average regularization scale. In this way, slope contributes not only to prediction but also to controlling representation adaptation.

Based on this insight, we propose \textbf{SlopeGuard}, a radial spectral slope-guided framework for generalizable AI-generated image detection. Slope-aware data processing diversifies training spectra and constructs fixed-size views from native-resolution patches. SlopeGuard estimates a compact slope descriptor from the luminance log-amplitude spectrum through stabilized closed-form regression and concatenates it with LoRA-adapted CLIP features for authenticity prediction. Batch-relative slope deviation further serves as an empirical proxy for weighting prior regularization. At inference, frequency diversification and the frozen prior branch are removed, while the adapted encoder and slope estimator process the reconstructed view for joint classification. Our main contributions are summarized as follows:
\begin{itemize}
\item We reveal generator-dependent heterogeneity in radial spectral slopes, showing that a universal threshold is unreliable and motivating slope as complementary evidence to learned visual representations.

\item We introduce SlopeGuard, a unified framework that integrates radial spectral slope with transferable visual representations for generalizable AI-generated image detection. The slope provides complementary evidence and guides prior regularization to suppress generator-specific representation drift.

\item Experiments demonstrate that SlopeGuard achieves state-of-the-art mean accuracies of 97.5\% on AIGCDetectBench, 97.8\% on GenImage, and 85.1\% on AIGIBench, while remaining robust to common image perturbations.
\end{itemize}

\begin{figure*}[t]
    \centering
    \includegraphics[width=\linewidth]{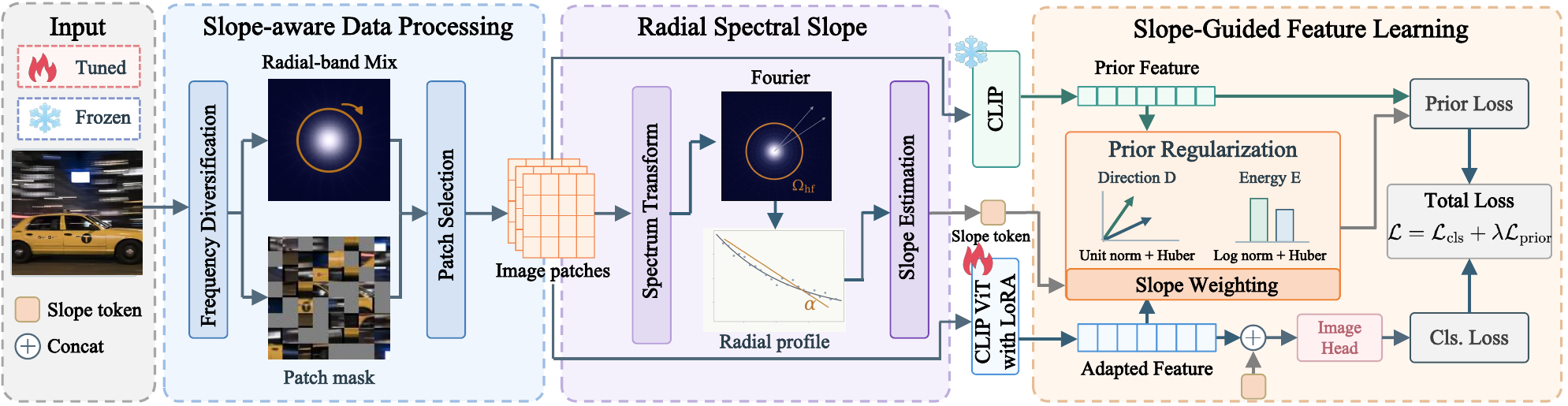}
\caption{
Overview of the proposed SlopeGuard framework.
Given input images, slope-aware data processing first diversifies frequency patterns, followed by Radial Spectral Slope estimation to extract a compact spectral signature.
The extracted slope cue is then integrated with CLIP features through slope-guided regularization to improve generalization across unseen generators.
}
    \label{fig:framework}
\end{figure*}

\section{Related Work}
\label{sec:formatting}

\subsection{AI-Generated Image Detection}
AI-generated image detection aims to distinguish real images from increasingly realistic content synthesized by generative models, including Generative Adversarial Networks (GANs) and diffusion models~\cite{corvi2023intriguing,huang2024gan,wu2024infinite,liu2026tsad}. Representative approaches span spatial-artifact analysis~\cite{yuan2026mlep,yang2026all,mu2026no}, frequency-aware modeling~\cite{chu2025fire}, and foundation-model adaptation~\cite{liu2024forgery,shi2025mirage}. Early spatial approaches employed convolutional neural networks or Vision Transformers to capture generator-specific artifacts, but they often generalized poorly to unseen generators~\cite{wang2020cnn,ojha2023towards}. Frequency-aware methods analyze Fourier spectra~\cite{tan2024frequency,yan2026dual}, wavelet-domain information, or learned frequency features~\cite{li2025improving} to capture systematic discrepancies between real and synthetic images. In parallel, foundation models such as CLIP have been adapted through prompt learning~\cite{li2026towards}, parameter-efficient fine-tuning~\cite{tan2025c2p,liu2026mirror}, and multimodal reasoning~\cite{zhou2025aigi}, showing promise for cross-generator detection.Despite this progress, robust generalization across unseen generators and synthesis pipelines remains challenging~\cite{zhang2025towards,zhang2025d}. Representative frequency-aware detectors construct discriminative representations from transformed spectra or learned frequency features~\cite{tan2024frequency,yan2026dual,li2025improving,li2026spectral}.

\subsection{Frequency-based AI-Generated Image Detection}
Frequency-domain representations have demonstrated strong robustness in AI-generated image detection, particularly under cross-generator evaluation~\cite{yan2026dual,tan2024frequency}. Existing methods leverage Fourier spectra, power spectral density, and joint spatial-frequency representations to characterize synthesis artifacts that are less dependent on semantic content~\cite{tan2023learning}. Despite their effectiveness, most approaches learn high-dimensional spectral representations while paying limited attention to the intrinsic statistical properties of frequency spectra~\cite{frank2020leveraging,durall2020watch}. Consequently, they tend to capture generator-specific frequency patterns rather than generator-invariant characteristics, limiting their generalization to unseen synthesis models~\cite{li2026reduce,karageorgiou2025any,tan2026aggregating}. To address this limitation, we introduce a regularization objective based on the Radial Spectral Slope, a compact statistical descriptor of spectral decay that encourages the model to learn more robust and transferable frequency representations for generalizable AI-generated image detection.

\section{Methodology}

\subsection{Overview of SlopeGuard}

As illustrated in Fig.~\ref{fig:framework}, SlopeGuard treats the Radial Spectral Slope as an explicit spectral cue complementary to the semantic representation of CLIP~\cite{radford2021learning}. The framework consists of three successive stages. Slope-aware data processing first constructs informative spectral views through frequency diversification and patch selection. Frequency diversification broadens the spectral patterns
observed during training, while patch selection constructs
fixed-size views without resizing individual patches,
retaining native-resolution detail within the selected regions. Radial Spectral Slope estimation then transforms each reconstructed view into the Fourier domain and fits the decay of its log-magnitude spectrum over a prescribed radial band, producing a compact slope estimate. Finally, slope-guided feature learning appends this estimate to the LoRA-adapted CLIP representation and uses it to guide prior regularization.


\subsection{Slope-Aware Data Processing}
Reliable slope-guided learning requires spectral evidence that is diverse enough to discourage generator-specific correlations yet sufficiently intact to retain localized synthesis traces. We therefore combine frequency diversification with patch selection. Frequency diversification perturbs radial decay and local frequency responses during training, whereas patch selection constructs fixed-size views that preserve spatially distributed spectral evidence.

\subsubsection{Frequency Diversification}
We stochastically augment real and generated training images before patch selection to diversify global spectral decay and local detail responses. Given an RGB image $\mathbf{x}$ with luminance $y$, we construct a luminance residual:
\begin{equation}
r =
\mathcal{F}^{-1}\!\left[
(G-\mathbf{1})\odot\mathcal{F}(y)
\right],
\label{eq:frequency_diversification}
\end{equation}
where $\mathcal{F}$ and $\mathcal{F}^{-1}$ denote the two-dimensional Fourier transform and its inverse, $\odot$ denotes elementwise multiplication, and $\mathbf{1}$ is an all-ones array. The positive radial gain map $G$ equals one outside a randomly jittered, resolution-scaled frequency band. Within the band, its logarithm varies linearly with normalized radius, with a randomly sampled positive or negative slope. This reshapes the radial amplitude profile while preserving phase at the spectral modulation stage. We broadcast $r$ across RGB channels, blend it into $\mathbf{x}$ with random strength, and clip the result to the valid intensity range.

\subsubsection{Patch Selection}
Following NS-Net~\cite{yan2025ns}, we extract nonoverlapping $32\times32$ patches at the original image resolution and rank them by normalized Fourier power entropy. During training, the $24$ lowest-entropy and $25$ highest-entropy patches are randomly arranged into a $7\times7$ grid. At inference, we combine central patches, spatial neighbors considered in random order, and both entropy extremes, fill remaining slots with near-median-entropy patches, and order the selected patches by their original coordinates. Both stages yield a $224\times224$ view without patch resizing; images with fewer than $49$ patches retain all patches with zero-filled unused slots. Subsequent spectral transformation and slope estimation operate on this reconstructed view, so RSS characterizes the radial spectral decay of the input shared with the CLIP encoder.

\subsection{Radial Spectral Slope}

The preceding stage yields fixed-size views containing spatially distributed spectral evidence, with frequency diversification applied only during training. We summarize each view using a parameter-free Radial Spectral Slope estimator. Here, slope-guided learning refers to learning representations conditioned on the estimated slope rather than learning the estimator itself. The resulting scalar is concatenated with the adapted CLIP representation as a slope token and, during training, further modulates prior regularization.

\subsubsection{Spectrum Transform}

Let $\mathbf{x}_b$ denote the reconstructed image of the $b$-th sample. We first reverse input normalization, clamp its values to $[0,1]$, and convert it to luminance $\mathbf{y}_b$. Exploiting the conjugate symmetry of real-valued inputs, we compute the orthonormal one-sided Fourier spectrum and obtain
\begin{equation}
\boldsymbol{\ell}_b
=
\log\!\left(
\left|\operatorname{rFFT}_{2}(\mathbf{y}_b)\right|
+\epsilon
\right).
\label{eq:spectrum_transform}
\end{equation}
Here, $\boldsymbol{\ell}_b$ denotes the log-magnitude spectrum and $\epsilon$ ensures numerical stability. Each coefficient $(u,v)$ is paired with radial frequency $\rho_{u,v}=(f_u^2+f_v^2)^{1/2}$, where $f_u$ and $f_v$ are the corresponding discrete Fourier coordinates.

\subsubsection{Slope Estimation}

The spectrum transform provides radial frequencies and their log-magnitude responses. We summarize their radial decay with a scalar slope. Given a reference band $[\rho_{\min},\rho_{\max}]$, we scale its boundaries by $s=R_{h,w}/R_{\mathrm{ref}}$, where $R_{h,w}$ and $R_{\mathrm{ref}}$ denote the maximum spectral radii at the input and reference resolutions. The adjusted band is $[s\rho_{\min},s\min(\rho_{\max},R_{\mathrm{ref}})]$, maintaining a consistent relative frequency range within valid spectral support. The reference band is specified separately from the frequency augmentation band.

Let $\Omega_{h,w}$ denote the coefficients whose radii fall within the adjusted band. We regress their log-magnitude responses directly against their radii. With $\widetilde{\rho}_i=\rho_i-\bar{\rho}$ and $\widetilde{\ell}_{b,i}=\ell_{b,i}-\bar{\ell}_b$ centered over $\Omega_{h,w}$, the stabilized least-squares slope is
\begin{equation}
\alpha_b
=
\frac{
\left\langle
\widetilde{\rho}_i\,
\widetilde{\ell}_{b,i}
\right\rangle_{\Omega_{h,w}}
}{
\left\langle
\widetilde{\rho}_i^{\,2}
\right\rangle_{\Omega_{h,w}}
+\epsilon
}.
\label{eq:slope_estimation}
\end{equation}
Here, $\langle\cdot\rangle_{\Omega_{h,w}}$ denotes averaging over the retained coefficients, and $\epsilon$ stabilizes the denominator. A more negative $\alpha_b$ indicates steeper spectral decay, whereas a larger value corresponds to a flatter or increasing radial trend. This scalar descriptor is integrated with the CLIP representation to guide feature learning.

\subsection{Slope-Guided Feature Learning}

The preceding spectrum transform and slope estimation convert each processed image into an explicit descriptor of radial spectral decay. Estimating the slope alone, however, does not determine how this spectral cue should interact with the semantic representation. Directly feeding a scalar slope to the classifier may encourage reliance on generator-dependent decision boundaries, while the adapted image encoder remains susceptible to artifacts associated with the training generators. This limitation becomes more pronounced for unseen generators whose spectral decay and visual content differ from those observed during training. We introduce slope-guided feature learning, which incorporates the estimated slope into authenticity prediction and uses it to modulate prior preservation. The resulting framework couples radial spectral behavior with transferable CLIP features through slope token integration and slope-guided prior regularization.

\subsubsection{Slope Token Integration}

Given an input image $\mathbf{x}_b$, the LoRA-adapted CLIP image encoder
$E_{\mathrm{img}}(\cdot)$ extracts a semantic representation
$\mathbf{z}_b\in\mathbb{R}^{d}$. Let $h_{\mathrm{img}}(\cdot)$ denote the
image classification head. We append the estimated radial spectral slope
$\alpha_b$ to the CLIP representation and feed the resulting
slope-conditioned feature to the classifier:

\begin{equation}
\mathbf{z}_b = E_{\mathrm{img}}(\mathbf{x}_b),\qquad
\widetilde{\mathbf{z}}_b=[\mathbf{z}_b;\alpha_b],\qquad
\widehat{y}_b=h_{\mathrm{img}}(\widetilde{\mathbf{z}}_b).
\label{eq:slope_token}
\end{equation}

Here, $E_{\mathrm{img}}$ denotes the LoRA-adapted CLIP image encoder,
$h_{\mathrm{img}}$ is the image classification head, and
$[\cdot\,;\cdot]$ denotes feature concatenation. This late integration
preserves the semantic representation produced by CLIP while providing the
classifier with an explicit radial spectral cue. It enables the prediction
head to exploit complementary spectral information without modifying the
CLIP token sequence.

\begin{table*}[h]
\centering
\caption{Comparison with state-of-the-art methods on the AIGCDetectBench~\cite{zhong2023patchcraft} using mean accuracy (\%). The best and second-best results are highlighted in \textbf{bold} and \underline{underlined}, respectively.
All other results are obtained using the official AIGIBench checkpoint~\cite{li2026artificial}. All models follow the same AIGIBench  training protocol~\cite{li2026artificial}.}
\label{tab:comparison_aigcacc}
\scriptsize
\setlength{\tabcolsep}{4pt}
\renewcommand{\arraystretch}{1}

\begin{tabular}{l ccccccc ccccccccc c|c}
\toprule
\multirow{3}{*}{\textbf{Method}}
& \multicolumn{7}{c}{\textit{GAN-based}}
& \multicolumn{9}{c}{\textit{Diffusion-based}}
& \multicolumn{1}{c}{\textit{Other}}
& \multirow{3}{*}{\textbf{mAcc.}} \\
\cmidrule(lr){2-8}
\cmidrule(lr){9-17}
\cmidrule(lr){18-18}
& \makecell{Big-\\GAN}
& \makecell{Pro-\\GAN}
& \makecell{Star-\\GAN}
& \makecell{Style-\\GAN}
& \makecell{Style-\\GAN2}
& \makecell{Cycle-\\GAN}
& \makecell{Gau-\\GAN}
& \makecell{Mid-\\journey}
& \makecell{Wu-\\kong}
& DALLE2 & SD1.5 & SD1.4 & SDXL & ADM & VQDM & GLIDE & WFIR & \\
\midrule

\textbf{CNNSpot}
& 50.6 & 97.6 & 60.5 & 68.2 & 61.0 & 59.2 & 51.2 & 52.5 & 63.7
& 52.1 & 77.7 & 77.7 & 51.8 & 50.7 & 50.2 & 52.3 & 50.3 & 60.4 \\

\textbf{Gram-Net}
& 54.0 & 98.5 & 94.8 & 75.4 & 69.3 & 56.9 & 50.0 & 67.1 & 96.4
& 66.8 & 99.2 & 99.4 & 72.4 & 57.3 & 53.3 & 76.4 & 46.7 & 72.6 \\

\textbf{UniFD}
& 85.9 & 98.4 & 97.5 & 75.4 & 72.7 & 88.0 & 92.1 & 78.3 & 90.4
& 78.1 & 95.7 & 96.1 & 84.9 & 53.9 & 70.8 & 85.5 & 70.4 & 83.2 \\

\textbf{NPR}
& 57.9 & 99.4 & 53.3 & 83.6 & 79.3 & 64.6 & 50.7 & 85.1 & 98.5
& 94.5 & 99.6 & 99.7 & 90.6 & 63.0 & 57.4 & 84.3 & 55.2 & 77.5 \\

\textbf{FreqNet}
& 72.7 & 99.3 & 94.2 & 86.9 & 83.3 & 78.7 & 51.8 & 86.4 & 94.7
& 82.0 & 95.6 & 95.6 & 90.8 & 64.9 & 62.9 & 85.4 & 61.5 & 81.6 \\

\textbf{SAFE}
& 85.4 & \textbf{100.0} & 100.0 & 89.3 & 97.9 & 95.3 & 85.0
& 87.3 & 99.4 & \underline{98.2} & \underline{99.8} & \underline{99.9}
& 99.5 & 64.6 & 91.4 & 94.4 & 52.1 & 90.5 \\

\textbf{AIDE}
& 75.9 & 97.2 & 97.8 & 94.5 & 98.2 & 92.9 & 63.8
& 81.5 & 99.2 & \textbf{99.3} & 99.6 & 99.6
& 97.6 & 95.9 & 94.8 & \underline{99.1} & 70.7 & 91.6 \\

\textbf{Effort}
& \textbf{99.2} & 100.0 & 100.0 & 89.8 & 91.0
& \textbf{99.9} & \textbf{99.6} & 72.1 & \textbf{99.9} & 87.9
& \textbf{99.9} & \textbf{100.0} & 89.8 & 76.5
& \underline{98.1} & 90.6 & \underline{95.5} & 93.5 \\

\textbf{DFFreq}
& 87.1 & 99.8 & 94.9 & \textbf{99.8} & \textbf{99.9}
& 97.9 & 85.1 & 94.7 & 93.5 & 96.8
& 94.5 & 94.2 & 98.1 & 94.2 & 96.5 & 95.2 & 52.2 & 92.6 \\

\textbf{BSF}
& 82.2 & 99.2 & \underline{99.7} & \underline{99.3} & 99.1
& 95.0 & 78.6 & \underline{98.7} & 99.1 & 99.3
& 99.2 & 99.1 & \underline{99.6} & \textbf{96.2}
& 97.3 & \textbf{99.3} & 52.1 & 93.7 \\

\textbf{LTD}
& \underline{97.4} & \underline{99.9} & 100.0
& 87.9 & 88.0 & \underline{99.6} & \underline{96.1}
& 77.9 & \underline{99.8} & 97.3
& 99.7 & 99.9 & 80.5 & 82.6
& 96.2 & 95.7 & \textbf{98.4} & \underline{93.9} \\

\midrule

\rowcolor{blue!10}\textbf{Ours}
& 91.5 & 99.8 & \textbf{100.0}
& 98.9 & \underline{99.7} & 98.7 & 86.7
& \textbf{98.9} & 99.7 & 98.0
& 99.6 & 99.7 & \textbf{99.9}
& \underline{96.0} & \textbf{99.2}
& 97.0 & 94.2 & \textbf{97.5} \\

\bottomrule
\end{tabular}
\end{table*}

\begin{table*}[htbp]
\centering
\caption{
Comparison with state-of-the-art methods on the GenImage benchmark~\cite{zhu2023genimage}
in terms of mean accuracy (\%) and average precision (\%).
The best and second-best results are highlighted in \textbf{bold} and
\underline{underlined}, respectively.
All models follow the same AIGIBench  training protocol~\cite{li2026artificial}.
}
\scriptsize
\setlength{\tabcolsep}{6pt}
\renewcommand{\arraystretch}{1}

\begin{tabular}{l cc cc cc cc cc cc cc cc |cc}
\toprule

\multirow{2}{*}{\textbf{Method}} &
\multicolumn{2}{c}{Midjourney} &
\multicolumn{2}{c}{SDv1.4} &
\multicolumn{2}{c}{SDv1.5} &
\multicolumn{2}{c}{ADM} &
\multicolumn{2}{c}{GLIDE} &
\multicolumn{2}{c}{Wukong} &
\multicolumn{2}{c}{VQDM} &
\multicolumn{2}{c}{BigGAN} &
\multicolumn{2}{c}{Mean} \\

\cmidrule(lr){2-3}
\cmidrule(lr){4-5}
\cmidrule(lr){6-7}
\cmidrule(lr){8-9}
\cmidrule(lr){10-11}
\cmidrule(lr){12-13}
\cmidrule(lr){14-15}
\cmidrule(lr){16-17}
\cmidrule(lr){18-19}

& Acc. & AP.
& Acc. & AP.
& Acc. & AP.
& Acc. & AP.
& Acc. & AP.
& Acc. & AP.
& Acc. & AP.
& Acc. & AP.
& Acc. & AP. \\

\midrule

\textbf{CNNSpot}
& 52.5 & 80.4
& 77.7 & 97.6
& 77.7 & 97.6
& 50.7 & 66.6
& 52.3 & 83.1
& 63.7 & 88.9
& 50.2 & 57.0
& 50.0 & 64.5
& 59.3 & 79.5 \\

\textbf{Gram-Net}
& 67.1 & 90.4
& 99.4 & \textbf{100.0}
& 99.2 & \underline{99.9}
& 57.3 & 71.5
& 76.4 & 94.6
& 96.4 & 99.7
& 53.3 & 61.5
& 60.9 & 90.3
& 76.2 & 88.5 \\

\textbf{LGrad}
& 80.7 & 91.2
& 95.7 & 99.3
& 95.6 & 99.2
& 60.0 & 68.5
& 88.2 & 96.0
& 92.7 & 97.9
& 59.8 & 69.4
& \underline{94.6} & 98.6
& 83.3 & 90.0 \\

\textbf{UniFD}
& 78.3 & 91.7
& 96.1 & 99.3
& 95.7 & 99.1
& 53.9 & 65.2
& 85.5 & 95.3
& 90.4 & 97.4
& 70.8 & 87.2
& 83.1 & 94.4
& 81.7 & 91.2 \\

\textbf{FreqNet}
& 86.4 & 94.5
& 95.7 & 99.9
& 95.6 & 99.9
& 64.6 & 74.1
& 85.3 & 94.1
& 94.7 & 99.1
& 62.6 & 74.9
& 94.6 & 99.3
& 84.9 & 92.0 \\

\textbf{NPR}
& 85.1 & 97.8
& 99.7 & 100.0
& 99.6 & 99.9
& 63.0 & 73.9
& 84.3 & 97.4
& 98.5 & \underline{99.9}
& 57.4 & 81.2
& 85.2 & 98.3
& 84.1 & 93.5 \\

\textbf{Effort}
& 72.1 & 98.9
& \textbf{100.0} & 100.0
& \textbf{99.9} & \textbf{100.0}
& 76.5 & 97.5
& 90.6 & \underline{99.8}
& \textbf{99.9} & 100.0
& \underline{98.1} & \textbf{100.0}
& \textbf{96.2} & \textbf{100.0}
& 91.7 & 99.5 \\

\textbf{SAFE}
& 87.3 & \underline{99.7}
& \underline{99.9} & 100.0
& \underline{99.8} & 100.0
& 64.6 & 89.1
& 94.4 & 99.1
& 99.4 & 99.9
& 91.4 & 99.5
& 91.5 & \underline{99.6}
& 91.0 & 98.4 \\

\textbf{AIDE}
& 81.5 & 90.3
& 99.6 & 99.9
& 99.6 & 99.9
& \underline{95.9} & \underline{99.3}
& \textbf{99.1} & \textbf{99.9}
& 99.2 & 99.9
& 94.8 & 98.8
& 86.6 & 98.2
& \underline{94.5} & 98.3 \\

\textbf{DFFreq}
& \underline{94.7} & 97.9
& 94.3 & 97.6
& 94.5 & 98.0
& 94.3 & 97.4
& 95.1 & 98.7
& 93.4 & 96.9
& 96.5 & 98.3
& 47.7 & 75.0
& 88.8 & 95.0 \\

\textbf{LTD}
& 77.9 & 98.8
& 99.9 & 100.0
& 99.7 & 100.0
& 82.7 & 99.2
& 95.7 & 99.8
& \underline{99.8} & 100.0
& 96.2 & 99.9
& 77.5 & 99.3
& 91.2 & \underline{99.6} \\

\midrule

\rowcolor{blue!10}
\textbf{Ours}
& \textbf{99.0} & \textbf{99.9}
& 99.7 & \underline{99.9}
& 99.6 & 99.8
& \textbf{96.0} & \textbf{99.5}
& \underline{96.9} & 99.6
& 99.7 & \textbf{100.0}
& \textbf{99.3} & \underline{99.9}
& 92.4 & 98.9
& \textbf{97.8} & \textbf{99.7} \\

\bottomrule
\end{tabular}

\label{tab:genimage}
\end{table*}

\subsubsection{Slope-Guided Prior Regularization}
Prior regularization constrains representation drift relative to the frozen CLIP prior. As shown in Fig.~\ref{fig:slope_distribution}, radial spectral slopes vary substantially across generators and between real and generated images. Motivated by this heterogeneity, we use standardized slope deviation as an empirical proxy for allocating prior regularization strength:
\begin{equation}
r_b=
\frac{|\alpha_b-\mu_{\alpha}|}
{\sigma_{\alpha}+\epsilon},
\qquad
w_b=
\frac{B\,\varphi(r_b)}
{\sum_{j=1}^{B}\varphi(r_j)}.
\label{eq:slope_weight}
\end{equation}
Here, $B$ is the mini-batch size, $\alpha_b$ denotes the radial spectral slope of sample $b$, and $\mu_{\alpha}$ and $\sigma_{\alpha}$ are the corresponding mini-batch mean and standard deviation. $r_b$ measures the standardized slope deviation, while $w_b$ is the resulting slope-guided regularization weight. We set $\varphi(r)=\tfrac{1}{2}+\operatorname{sigmoid}(r)$, a bounded, strictly increasing mapping, and $\epsilon$ is a small constant for numerical stability. The normalization ensures $\frac{1}{B}\sum_{b=1}^{B}w_b=1$, maintaining a stable average regularization strength.

Let $\mathbf{a}_b,\mathbf{p}_b\in\mathbb{R}^{d}$ denote the adapted and prior features, respectively, where $d$ is the feature dimension. We preserve both feature direction and energy. Define
\begin{equation}
\widehat{\mathbf{v}}
=
\frac{\mathbf{v}}
{\|\mathbf{v}\|_2+\epsilon},
\qquad
e(\mathbf{v})
=
\log(1+\|\mathbf{v}\|_2).
\label{eq:feature_transform}
\end{equation}
The directional consistency term is
\begin{equation}
\mathcal{D}_b=
\frac{1}{d}
\sum_{i=1}^{d}
h\!\left(
\widehat{a}_{b,i}-\widehat{p}_{b,i}
\right),
\label{eq:direction_consistency}
\end{equation}
while the energy consistency term is
\begin{equation}
\mathcal{E}_b=
h\!\left(
e(\mathbf{a}_b)-e(\mathbf{p}_b)
\right),
\label{eq:energy_consistency}
\end{equation}
where $h(\cdot)$ denotes the unit-threshold Huber penalty. Thus, $\mathcal{D}_b$ constrains directional deviation, whereas $\mathcal{E}_b$ preserves feature magnitude.

The slope-guided prior regularization objective is
\begin{equation}
\mathcal{L}_{\mathrm{prior}}
=
\frac{1}{B}
\sum_{b=1}^{B}
w_b
\left(
\mathcal{D}_b+
\frac{1}{2}\mathcal{E}_b
\right).
\label{eq:prior_regularization}
\end{equation}
Consequently, spectrally atypical samples receive stronger prior preservation, whereas samples near the batch-central slope retain greater adaptation flexibility.
\subsubsection{Training Objective}

We jointly optimize forensic discrimination and prior preservation by combining the classification loss with the slope-guided prior objective
\begin{equation}
\mathcal{L}
=
\mathcal{L}_{\mathrm{cls}}
+
\lambda_{\mathrm{prior}}
\mathcal{L}_{\mathrm{prior}},
\label{eq:training_objective}
\end{equation}
where $\mathcal{L}_{\mathrm{cls}}$ drives authenticity prediction and $\lambda_{\mathrm{prior}}$ controls the strength of prior preservation. The hyperparameter $\lambda_{\mathrm{prior}}$ is set to 0.1. This joint objective incorporates radial slope into both feature integration and prior regularization, encouraging discriminative adaptation without excessive deviation from the transferable CLIP representation.

\section{Experiments}

\begin{table*}[h]
\centering
\caption{Comparison with state-of-the-art methods on AIGIBench~\cite{li2026artificial}, evaluated using mAcc and AP.  $^{\clubsuit}$ Results obtained using the official AIGIBench checkpoint~\cite{li2026artificial}. $^{\spadesuit}$ Results obtained using the official DGS-Net checkpoint~\cite{yan2025dgs}. All other results are reported in the original AIGIBench paper~\cite{li2026artificial}. All models follow the same AIGIBench  training protocol~\cite{li2026artificial}.}
\label{tab:aigi}

\begin{threeparttable}
\resizebox{\textwidth}{!}{%
\setlength{\tabcolsep}{3.1pt}
\renewcommand{\arraystretch}{1}

\begin{tabular}{
l|cc|cc|cc|cc|cc|cc|cc|cc|cc|cc|cc|
>{\columncolor{blue!10}}c
>{\columncolor{blue!10}}c
}
\toprule

\multirow{2}{*}{\textbf{Method}} &
\multicolumn{2}{c}{\textbf{CNNSpot}} &
\multicolumn{2}{c}{\textbf{Gram-net}} &
\multicolumn{2}{c}{\textbf{NPR}} &
\multicolumn{2}{c}{\textbf{FreqNet}} &
\multicolumn{2}{c}{\textbf{LGrad}} &
\multicolumn{2}{c}{\textbf{BSF$^{\clubsuit}$}} &
\multicolumn{2}{c}{\textbf{SAFE}} &
\multicolumn{2}{c}{\textbf{AIDE}} &
\multicolumn{2}{c}{\textbf{DFFreq}} &
\multicolumn{2}{c}{\textbf{Effort$^{\clubsuit}$}} &
\multicolumn{2}{c}{\textbf{DGS-Net$^{\spadesuit}$}} &
\multicolumn{2}{>{\columncolor{blue!10}}c}{\textbf{Ours}} \\

\cmidrule(lr){2-3}
\cmidrule(lr){4-5}
\cmidrule(lr){6-7}
\cmidrule(lr){8-9}
\cmidrule(lr){10-11}
\cmidrule(lr){12-13}
\cmidrule(lr){14-15}
\cmidrule(lr){16-17}
\cmidrule(lr){18-19}
\cmidrule(lr){20-21}
\cmidrule(lr){22-23}
\cmidrule(lr){24-25}

& \footnotesize Acc. & \footnotesize AP.
& \footnotesize Acc. & \footnotesize AP.
& \footnotesize Acc. & \footnotesize AP.
& \footnotesize Acc. & \footnotesize AP.
& \footnotesize Acc. & \footnotesize AP.
& \footnotesize Acc. & \footnotesize AP.
& \footnotesize Acc. & \footnotesize AP.
& \footnotesize Acc. & \footnotesize AP.
& \footnotesize Acc. & \footnotesize AP.
& \footnotesize Acc. & \footnotesize AP.
& \footnotesize Acc. & \footnotesize AP.
& \footnotesize Acc. & \footnotesize AP. \\

\midrule

ProGAN
& 97.6 & 99.9
& 98.5 & 100.0
& 99.4 & 100.0
& 99.3 & 100.0
& 96.6 & 99.8
& 99.3 & 100.0
& \underline{100.0} & 100.0
& 97.2 & 99.6
& 98.1 & 100.0
& \textbf{100.0} & 100.0
& 99.7 & \underline{100.0}
& 99.8 & \textbf{100.0} \\

R3GAN
& 50.4 & 52.7
& 47.9 & 52.5
& 50.8 & 61.1
& 62.3 & 56.8
& 54.4 & 58.7
& \underline{94.7} & 93.9
& 93.9 & \underline{98.2}
& 92.9 & 97.1
& 61.7 & 74.0
& \textbf{95.3} & \textbf{98.2}
& 90.5 & 97.0
& 93.8 & 97.6 \\

StyleGAN3
& 55.8 & 73.1
& 65.6 & 77.9
& 78.4 & 91.7
& 83.0 & 92.4
& 70.5 & 80.5
& 93.4 & 95.8
& 89.7 & 97.6
& 88.1 & 91.4
& 90.1 & 96.4
& \textbf{95.5} & \textbf{98.9}
& 90.7 & \underline{98.3}
& \underline{94.2} & 97.7 \\

StyleGAN-XL
& 52.8 & 64.2
& 72.9 & 83.7
& 60.3 & 75.3
& 79.8 & 84.1
& 65.7 & 74.6
& \textbf{94.3} & 97.2
& 93.1 & \underline{97.6}
& 88.7 & 93.2
& 73.7 & 83.4
& 89.6 & 97.0
& 89.6 & \textbf{97.8}
& \underline{93.3} & 96.7 \\

StyleSwim
& 52.6 & 76.5
& 75.5 & 86.0
& 85.7 & 94.9
& 80.8 & 91.8
& 81.3 & 90.0
& 94.6 & 98.6
& \textbf{97.8} & \underline{99.6}
& 83.7 & 89.3
& 84.5 & 92.9
& \underline{96.8} & 99.1
& 89.8 & \textbf{99.6}
& 93.3 & 95.7 \\

WFIR
& 49.8 & 50.0
& 44.5 & 43.9
& 51.6 & 65.5
& 58.5 & 48.9
& 51.7 & 49.4
& 52.1 & 51.1
& 60.4 & 81.8
& 71.4 & 90.8
& 74.6 & 82.2
& \textbf{95.5} & \textbf{100.0}
& 92.3 & 99.8
& \underline{94.8} & \underline{99.9} \\

BlendFace
& 52.4 & \textbf{73.4}
& 42.3 & 33.3
& 44.5 & 34.7
& 23.3 & 34.1
& 41.8 & 34.9
& 46.2 & 47.9
& 47.3 & 45.6
& 51.5 & 54.2
& 41.5 & 35.2
& 52.7 & 66.1
& \underline{68.1} & 65.4
& \textbf{73.8} & \underline{69.8} \\

E4S
& 51.1 & \underline{68.9}
& 42.5 & 32.5
& 45.0 & 34.4
& 25.8 & 34.7
& 41.5 & 32.8
& 42.2 & 38.4
& 47.6 & 46.0
& 44.3 & 44.3
& 42.1 & 34.8
& 56.4 & \textbf{76.9}
& \underline{56.9} & 56.7
& \textbf{57.6} & 62.7 \\

FaceSwap
& 50.3 & 58.7
& 47.0 & 34.6
& 48.1 & 43.6
& 40.4 & 43.4
& 45.3 & 37.5
& 52.1 & 66.3
& 50.7 & 45.7
& 52.1 & 56.3
& 47.6 & 45.6
& 56.1 & \textbf{85.8}
& \textbf{76.4} & 79.6
& \underline{74.5} & \underline{81.0} \\

InSwap
& 54.5 & 77.9
& 47.1 & 38.0
& 47.8 & 40.7
& 37.5 & 42.1
& 44.6 & 35.0
& 48.1 & 49.5
& 49.7 & 49.9
& 50.9 & 54.6
& 47.0 & 41.2
& 58.0 & \underline{83.1}
& \underline{78.7} & 82.8
& \textbf{81.9} & \textbf{83.8} \\

SimSwap
& 52.1 & 70.0
& 46.7 & 36.4
& 47.4 & 42.7
& 36.5 & 41.9
& 44.2 & 37.6
& 51.2 & 63.0
& 49.0 & 49.5
& 54.9 & 62.7
& 45.6 & 43.8
& 66.5 & \textbf{89.7}
& \underline{83.8} & \underline{86.1}
& \textbf{86.1} & 85.3 \\

FLUX1-dev
& 57.4 & 72.3
& 64.5 & 75.1
& \underline{95.2} & \underline{99.0}
& 78.5 & 87.3
& 80.4 & 88.1
& 94.3 & 97.4
& \textbf{98.1} & \textbf{99.7}
& 88.0 & 93.4
& 76.6 & 86.3
& 61.4 & 74.4
& 88.9 & 96.6
& 93.1 & 96.6 \\

Midjourney
& 52.3 & 59.8
& 43.9 & 41.6
& 68.8 & 76.9
& 53.9 & 55.9
& 60.4 & 64.5
& \underline{84.6} & 86.2
& \textbf{94.1} & \textbf{98.4}
& 76.4 & 83.0
& 64.6 & 68.1
& 80.3 & 90.0
& 73.0 & 93.0
& 84.3 & \underline{94.0} \\

GLIDE
& 51.1 & 60.0
& 71.6 & 84.3
& 82.5 & 94.3
& 75.8 & 77.4
& 82.4 & 91.0
& \textbf{96.5} & \textbf{98.4}
& 92.5 & \underline{97.9}
& \underline{93.4} & 97.7
& 88.5 & 96.3
& 89.1 & 96.5
& 92.5 & 97.4
& 92.5 & 93.9 \\

DALLE-3
& 53.9 & \underline{68.6}
& 53.5 & 61.1
& 57.1 & \textbf{70.0}
& \textbf{66.2} & 61.0
& \underline{57.2} & 62.7
& 45.8 & 43.0
& 49.0 & 45.8
& 55.1 & 63.1
& 51.8 & 58.6
& 54.8 & 60.6
& 41.5 & 40.8
& 45.5 & 52.8 \\

Imagen3
& 51.2 & 57.4
& 50.6 & 54.7
& 85.9 & 94.4
& 73.6 & 80.7
& 62.2 & 69.7
& 93.4 & 94.1
& \textbf{96.7} & \textbf{98.8}
& 89.8 & 95.2
& 75.7 & 87.3
& 90.7 & \underline{97.8}
& 89.9 & 94.3
& \underline{93.9} & 97.3 \\

SD3
& 55.8 & 73.1
& 52.5 & 60.0
& 91.9 & 97.2
& 77.3 & 82.6
& 63.4 & 72.1
& \textbf{95.5} & 96.4
& 94.1 & \underline{98.8}
& \underline{94.3} & 98.3
& 81.3 & 90.8
& 92.1 & 97.9
& 91.8 & \textbf{99.0}
& 94.2 & 98.6 \\

SDXL
& 52.8 & 64.2
& 63.8 & 76.9
& 86.6 & 94.4
& 82.7 & 95.2
& 73.6 & 82.8
& \underline{96.2} & 98.0
& \textbf{98.3} & \textbf{99.7}
& 93.5 & 95.7
& 88.9 & 95.8
& 91.8 & 98.0
& 92.0 & \underline{99.1}
& 94.4 & 98.1 \\

BLIP
& 77.2 & 92.9
& 98.6 & 99.9
& 99.2 & 100.0
& 93.8 & 100.0
& 93.0 & 97.4
& \textbf{99.9} & 100.0
& 99.7 & 100.0
& 96.4 & 95.5
& 97.9 & 99.6
& \underline{99.7} & \underline{100.0}
& 98.5 & \textbf{100.0}
& 98.5 & 99.1 \\

Infinite-ID
& 49.7 & 49.5
& 50.6 & 60.1
& 63.9 & 80.4
& 79.0 & 74.5
& 50.9 & 54.6
& \underline{95.5} & 97.0
& \textbf{96.9} & \textbf{99.2}
& 92.2 & 94.7
& 70.4 & 82.7
& 69.8 & 85.7
& 90.7 & \underline{97.8}
& 93.5 & 94.9 \\

InstantID
& 53.2 & 80.2
& 75.1 & 85.7
& 63.8 & 79.2
& 79.8 & 86.3
& 72.6 & 81.5
& \underline{94.8} & \underline{99.0}
& \textbf{98.2} & \textbf{99.6}
& 91.8 & 96.3
& 91.9 & 97.2
& 71.2 & 87.1
& 89.5 & 97.3
& 93.2 & 91.7 \\

IP-Adapter
& 52.0 & 65.8
& 54.5 & 64.2
& 82.4 & 91.7
& 78.8 & 79.9
& 70.3 & 78.3
& \textbf{95.3} & \underline{98.7}
& 92.8 & 98.1
& 90.0 & 95.4
& 83.2 & 91.3
& 92.5 & 98.1
& 90.2 & \textbf{99.1}
& \underline{93.8} & 95.1 \\

PhotoMaker
& 50.1 & 58.2
& 50.1 & 58.6
& 48.1 & 43.6
& 77.0 & 74.9
& 59.9 & 67.2
& \underline{95.3} & \underline{98.7}
& \textbf{97.0} & \textbf{99.3}
& 91.7 & 95.6
& 88.0 & 94.0
& 59.8 & 69.2
& 87.9 & 94.1
& 93.4 & 94.5 \\

SocialRF
& 51.1 & 50.6
& 52.1 & 53.0
& 59.1 & 68.4
& 54.2 & 58.1
& 53.5 & 54.9
& \textbf{60.7} & 64.1
& 58.0 & 64.2
& 57.8 & 65.0
& 57.6 & 63.3
& 58.1 & 65.2
& 59.3 & \underline{71.2}
& \underline{59.9} & \textbf{77.0} \\

CommunityAI
& 51.3 & 59.1
& 52.6 & 66.1
& 54.0 & 62.9
& \textbf{55.9} & \textbf{69.7}
& \underline{55.5} & \underline{69.4}
& 54.9 & 57.5
& 54.2 & 55.2
& 54.1 & 61.0
& 54.5 & 52.1
& 52.6 & 64.3
& 54.6 & 52.0
& 54.7 & 65.2 \\

\midrule

\textbf{Mean}
& 55.1 & 67.1
& 58.6 & 62.4
& 67.9 & 73.9
& 66.2 & 70.1
& 62.9 & 66.6
& 78.8 & 81.1
& 79.9 & 82.6
& 77.6 & 82.7
& 71.1 & 75.7
& 77.1 & 87.2
& \underline{82.3} & \underline{87.8}
& \textbf{85.1} & \textbf{88.8} \\

\bottomrule
\end{tabular}%
}
\end{threeparttable}
\end{table*}

\begin{table}[t]
\centering
\caption{Comparison with state-of-the-art methods on AIGIBench~\cite{li2026artificial}. $^{\spadesuit}$ Results obtained using the official NS-Net checkpoint~\cite{yan2025ns}. The other results of competing methods are taken from the official AIGIBench implementation and released results~\cite{li2026artificial}.}
\label{tab:aigicomparison}

\small
\setlength{\tabcolsep}{4.5pt}
\renewcommand{\arraystretch}{1}

\begin{tabular}{lccc}
\toprule
Method & Ref. & mAcc. & AP. \\
\midrule
C2P-CLIP$^{*}$       & AAAI'25    & 71.8 & 82.2 \\
VIB-Net              & CVPR'25    & 69.3 & 70.9 \\
D$^{3}$              & CVPR'25    & 63.7 & 68.9 \\
FerretNet            & NeurIPS'25 & \underline{79.4} & \underline{85.8} \\
LOTA                 & ICCV'25    & 77.4 & 83.1 \\
LTD                  & CVPR'26    & 74.9 & 77.6 \\
NS-Net$^{\spadesuit}$ & TMM'26     & 77.5 & 82.8 \\
\midrule
\rowcolor{blue!10}
\textbf{Ours}        & --         & \textbf{85.1} & \textbf{88.8} \\
\bottomrule
\end{tabular}
\end{table}

\subsection{Implementation Details}
Our model is implemented in PyTorch and trained on a single NVIDIA RTX 5090 GPU for one epoch. We use Adam~\cite{kingma2014adam} with an initial learning rate of $1\times10^{-4}$ and a batch size of 32. Following prior work~\cite{tan2025c2p,ojha2023towards}, we adopt pretrained CLIP ViT-L/14~\cite{radford2021learning} as the backbone and apply LoRA to the vision encoder with rank $r=6$, scaling factor $\alpha=6$, and dropout $0.8$. The classification threshold is fixed at 0.5 for all evaluations. We set the random seed to 100 for reproducibility. 

\subsection{Datasets and Metrics.}
Following the official protocol of AIGIBench~\cite{li2026artificial}, we train on 144K AI-generated images produced by SD-v1.4~\cite{rombach2022high} and ProGAN~\cite{karras2017progressive} across four object categories, along with an equal number of real images. For a fair comparison, all compared methods use the same training set.
To comprehensively evaluate the effectiveness and cross-dataset generalization of the methods, we conduct experiments on three representative AIGI detection benchmarks, including GenImage~\cite{zhu2023genimage}, AIGCDetectBench~\cite{zhong2023patchcraft}, and AIGIBench~\cite{li2026artificial}. 
 Following previous works~\cite{yan2025dgs}, we adopt Average Precision (AP.) and mean Accuracy (mAcc.) as the evaluation metrics.

\subsection{Compared Methods}

We compare nineteen AIGI detectors in three categories:
(1) artifact-based methods, including CNNSpot and Gram-Net (CVPR'20) \cite{wang2020cnn,liu2020global}, LGrad (CVPR'23) \cite{tan2023learning}, FreqNet (AAAI'24) \cite{tan2024frequency}, NPR (CVPR'24) \cite{tan2024rethinking}, SAFE (KDD'25) \cite{li2025improving}, LOTA (ICCV'25) \cite{wang2025lota}, FerretNet (NeurIPS'25) \cite{liang2026ferretnet}, DFFreq (TIFS'26) \cite{yan2026dual}, and BSF (AAAI'26) \cite{zhou2026beyond};
(2) methods using or adapting foundation-model features, including UniFD (CVPR'23) \cite{ojha2023towards}, C2P-CLIP$^{*}$ (AAAI'25) \cite{tan2025c2p}, VIB-Net (CVPR'25) \cite{zhang2025towards}, Effort (ICML'25) \cite{yan2025orthogonal}, LTD (CVPR'26) \cite{yang2026layer}, and DGS-Net (ICML'26) \cite{yan2025dgs};
and (3) hybrid methods combining both, including D$^{3}$ (CVPR'25) \cite{yang2025d}, AIDE (ICLR'25) \cite{yan2025sanity}, and NS-Net (TMM'26) \cite{yan2025ns}.

\subsection{Comparison with the State-of-the-Art}

\subsubsection{Evaluation on AIGCDetectBench}
As shown in Table~\ref{tab:comparison_aigcacc}, our method achieves the highest mAcc. of 97.5\% on AIGCDetectBench, surpassing LTD, DFFreq, and BSF by 3.6, 4.9, and 3.8 percentage points, respectively. It ranks first or second on multiple GAN- and diffusion-based generators, including Midjourney, SDXL, and VQDM. Moreover, it exceeds 90\% accuracy on 16 of the 17 generators, demonstrating consistently strong performance and cross-generator generalization across the diverse synthesis paradigms in this benchmark.

\subsubsection{Evaluation on GenImage}
As shown in Table~\ref{tab:genimage}, our method achieves the best overall performance on GenImage, attaining 97.8\% mAcc and 99.7\% mAP. These results surpass the strongest competing methods by 3.3 and 0.1 percentage points, respectively. Compared with AIDE, which achieves the second-best mAcc of 94.5\%, our method provides a substantial improvement in classification accuracy while maintaining strong ranking performance. Across individual generators, it achieves accuracies ranging from 92.4\% to 99.7\%, with AP consistently above 98\%. This consistently strong performance across diverse synthesis methods supports the effectiveness of our approach in cross-generator generalization.

\subsubsection{Evaluation on AIGIBench}
As shown in Tables~\ref{tab:aigi} and~\ref{tab:aigicomparison}, our method achieves the best overall performance on AIGIBench, reaching 85.1\% mAcc. and 88.8\% AP. Specifically, it surpasses FerretNet by 5.7 and 3.0 percentage points in mAcc. and AP, respectively, and further outperforms DGS-Net by 2.8 and 1.0 percentage points on the corresponding metrics. Beyond these aggregate improvements, our method also delivers competitive results across individual generators. Among the 25 generators evaluated, it ranks first on 4 and second on 7, achieving top-two performance on 11 generators in total. These results across diverse synthesis paradigms support the effectiveness of our method and its strong cross-generator generalization.

\begin{table*}[t]
\centering
\caption{Robustness evaluation on AIGIBench~\cite{li2026artificial} under different image perturbations. The last row applies all perturbations simultaneously. Best results are shown in bold. All other results are obtained using the official AIGIBench checkpoint~\cite{li2026artificial}.}
\label{tab:robustness1}

\setlength{\tabcolsep}{4pt}
\renewcommand{\arraystretch}{1}

\resizebox{\textwidth}{!}{%
\begin{tabular}{
cccc|cc|cc|cc|cc|cc|cc|cc|
>{\columncolor{blue!10}}c
>{\columncolor{blue!10}}c
}
\toprule

\multicolumn{4}{c|}{Perturbation}
& \multicolumn{2}{c|}{NPR}
& \multicolumn{2}{c|}{FreqNet}
& \multicolumn{2}{c|}{UniFD}
& \multicolumn{2}{c|}{AIDE}
& \multicolumn{2}{c|}{SAFE}
& \multicolumn{2}{c|}{DFFreq}
& \multicolumn{2}{c|}{BSF}
& \multicolumn{2}{>{\columncolor{blue!10}}c}{Ours} \\

\cmidrule(lr){1-4}
\cmidrule(lr){5-6}
\cmidrule(lr){7-8}
\cmidrule(lr){9-10}
\cmidrule(lr){11-12}
\cmidrule(lr){13-14}
\cmidrule(lr){15-16}
\cmidrule(lr){17-18}
\cmidrule(lr){19-20}

Blur & Crop & JPEG & Noise
& mAcc. & AP.
& mAcc. & AP.
& mAcc. & AP.
& mAcc. & AP.
& mAcc. & AP.
& mAcc. & AP.
& mAcc. & AP.
& mAcc. & AP. \\

\midrule

$\checkmark$ & $\times$ & $\times$ & $\times$
& 70.8 & 78.0
& 67.7 & 73.5
& 72.4 & 76.2
& 77.2 & \underline{81.7}
& \underline{79.6} & 74.1
& 78.6 & 77.7
& 79.1 & 76.9
& \textbf{80.8} & \textbf{84.2} \\

$\times$ & $\checkmark$ & $\times$ & $\times$
& 68.3 & 74.6
& 65.4 & 69.7
& 71.4 & 74.4
& 75.8 & 82.0
& 77.0 & 81.1
& 76.9 & 81.6
& \underline{78.3} & \underline{82.2}
& \textbf{83.2} & \textbf{88.4} \\

$\times$ & $\times$ & $\checkmark$ & $\times$
& 59.0 & 65.4
& 58.3 & 62.1
& \underline{66.4} & 71.4
& 63.9 & \underline{71.6}
& 64.8 & 69.2
& 64.1 & 70.4
& 64.6 & 67.6
& \textbf{67.2} & \textbf{75.3} \\

$\times$ & $\times$ & $\times$ & $\checkmark$
& 60.5 & 69.3
& 59.5 & 64.2
& 68.9 & 72.4
& 69.2 & 80.2
& 65.3 & 74.7
& \underline{78.2} & 78.8
& 71.5 & \underline{81.9}
& \textbf{81.7} & \textbf{84.3} \\

\midrule

$\checkmark$ & $\checkmark$ & $\checkmark$ & $\checkmark$
& 64.7 & 72.1
& 62.9 & 67.4
& 69.8 & 73.5
& 71.6 & \underline{79.5}
& 71.9 & 73.9
& \underline{74.5} & 77.1
& 73.5 & 76.8
& \textbf{78.2} & \textbf{83.0} \\

\bottomrule
\end{tabular}%
}

\end{table*}
\begin{table*}[h]
\centering
\caption{Robustness evaluation on AIGCDetectBench~\cite{zhong2023patchcraft} under different image perturbations. The last row applies all perturbations simultaneously. All other results are obtained using the official AIGIBench checkpoint~\cite{li2026artificial}.}
\label{tab:robustness2}

\setlength{\tabcolsep}{4pt}
\renewcommand{\arraystretch}{1}

\resizebox{\textwidth}{!}{%
\begin{tabular}{
cccc|cc|cc|cc|cc|cc|cc|cc|
>{\columncolor{blue!10}}c
>{\columncolor{blue!10}}c
}
\toprule

\multicolumn{4}{c|}{Perturbation}
& \multicolumn{2}{c|}{NPR}
& \multicolumn{2}{c|}{FreqNet}
& \multicolumn{2}{c|}{UniFD}
& \multicolumn{2}{c|}{AIDE}
& \multicolumn{2}{c|}{SAFE}
& \multicolumn{2}{c|}{DFFreq}
& \multicolumn{2}{c|}{BSF}
& \multicolumn{2}{>{\columncolor{blue!10}}c}{Ours} \\

\cmidrule(lr){1-4}
\cmidrule(lr){5-6}
\cmidrule(lr){7-8}
\cmidrule(lr){9-10}
\cmidrule(lr){11-12}
\cmidrule(lr){13-14}
\cmidrule(lr){15-16}
\cmidrule(lr){17-18}
\cmidrule(lr){19-20}

Blur & Crop & JPEG & Noise
& mAcc. & AP.& mAcc. & AP.& mAcc. & AP.& mAcc. & AP.& mAcc. & AP.
& mAcc. & AP.& mAcc. & AP.& mAcc. & AP. \\

\midrule

$\checkmark$ & $\times$ & $\times$ & $\times$
& 81.1 & 88.2
& 75.4 & 82.4
& 81.9 & 91.7
& 86.5 & \underline{92.7}
& 85.1 & 85.7
& 86.0 & 86.4
& \underline{86.6} & 84.6
& \textbf{90.7} & \textbf{97.0} \\

$\times$ & $\checkmark$ & $\times$ & $\times$
& 77.5 & 91.7
& 80.7 & 88.1
& 83.2 & 92.5
& 88.8 & 94.6
& 86.5 & 94.5
& 91.4 & 94.8
& \underline{93.2} & \underline{95.4}
& \textbf{95.7} & \textbf{99.2} \\

$\times$ & $\times$ & $\checkmark$ & $\times$
& 63.9 & 73.7
& 66.3 & 70.1
& 70.5 & 84.1
& 71.6 & \textbf{84.1}
& 70.3 & 80.9
& 71.7 & 77.1
& \underline{72.3} & 75.8
& \textbf{73.4} & \underline{83.2} \\

$\times$ & $\times$ & $\times$ & $\checkmark$
& 66.7 & 81.2
& 69.9 & 77.2
& 76.8 & 88.3
& 79.4 & \underline{91.6}
& 70.7 & 83.4
& \underline{84.0} & 88.5
& 82.3 & 85.2
& \textbf{87.6} & \textbf{95.3} \\

\midrule

$\checkmark$ & $\checkmark$ & $\checkmark$ & $\checkmark$
& 72.1 & 82.7
& 72.9 & 78.0
& 78.0 & 88.9
& 81.4 & \underline{90.6}
& 77.9 & 84.4
& \underline{83.3} & 85.7
& 77.5 & 89.3
& \textbf{86.7} & \textbf{93.5} \\
\bottomrule
\end{tabular}%
}
\end{table*}
\subsection{Ablation Study}


\subsubsection{Effect of each proposed component.}
Table~\ref{tab:component_ablation} shows the cumulative contribution of each component. Slope-aware Data Processing (SADP) improves mAcc.\ by 6.8 points. Adding Radial Spectral Slope (RSS) further improves mAcc.\ by 4.9 points, demonstrating the complementary value of spectral decay cues. Incorporating Slope-guided Prior Regularization (SGPR) brings an additional improvement of 1.4 points. Overall, the full model surpasses the baseline by 13.1 points in mAcc., highlighting the effectiveness of the proposed components.

\subsubsection{Effect of Design Choices}
Table~\ref{tab:design_choices} examines the slope source and
patch selection strategy.
Estimating slopes from reconstructed views achieves 85.1\%
mAcc., outperforming original images by 6.6 percentage points.
Patch shuffling improves mAcc.\ from 81.2\% to 85.1\%,
supporting its use in patch selection.


\definecolor{checkblue}{RGB}{36,105,170}
\definecolor{crossgray}{RGB}{95,95,95}
\definecolor{rowblue}{RGB}{239,245,251}

\newcommand{\cmark}{\textcolor{checkblue}{\ding{51}}}
\newcommand{\xmark}{\textcolor{crossgray}{\ding{55}}}

\begin{table}[t]
\centering
\begin{minipage}[t]{0.49\columnwidth}
\vspace{0pt}
\centering
\caption{Component ablation on AIGIBench~\cite{li2026artificial} (\%).}
\label{tab:component_ablation}

\scriptsize
\setlength{\tabcolsep}{3pt}
\renewcommand{\arraystretch}{1.15}

\def\ablationgap{%
  \dimexpr
  (\aboverulesep+\lightrulewidth+\belowrulesep)/5
  \relax
}

\begin{tabularx}{\linewidth}{*{4}{>{\centering\arraybackslash}X}}
\toprule
\textbf{SADP} & \textbf{RSS} & \textbf{SGPR} & \textbf{mAcc.}
\\[\ablationgap]
\midrule
\xmark & \xmark & \xmark & 72.0 \\[\ablationgap]
\cmark & \xmark & \xmark & 78.8 \\[\ablationgap]
\cmark & \xmark & \cmark & 79.2 \\[\ablationgap]
\cmark & \cmark & \xmark & 83.7 \\[\ablationgap]
\midrule
\rowcolor{blue!10}
\cmark & \cmark & \cmark & \textbf{85.1} \\
\bottomrule
\end{tabularx}
\end{minipage}\hfill%
\begin{minipage}[t]{0.49\columnwidth}
\vspace{0pt}
\centering
\caption{Design choices on AIGIBench~\cite{li2026artificial} (\%).}
\label{tab:design_choices}

\scriptsize
\setlength{\tabcolsep}{3pt}
\renewcommand{\arraystretch}{1.15}

\begin{tabularx}{\linewidth}{>{\raggedright\arraybackslash}X c}
\toprule
\textbf{Slope source} & \textbf{mAcc.} \\
\midrule
Original image & 78.5 \\
\rowcolor{blue!10}
Reconstructed view & \textbf{85.1} \\
\midrule
\textbf{Patch selection} & \textbf{mAcc.} \\
\midrule
w/o Shuffle & 81.2 \\
\rowcolor{blue!10}
w/ Shuffle & \textbf{85.1} \\
\bottomrule
\end{tabularx}
\end{minipage}
\end{table}

\begin{table}[t]
    \centering
    \begin{minipage}[t]{0.53\linewidth}
        \vspace{0pt}
        \centering
        \caption{Prior regularization on AIGIBench~\cite{li2026artificial} (\%).}
        \label{tab:prior_ablation}
        \small
        \setlength{\tabcolsep}{3pt}
        \renewcommand{\arraystretch}{1}
        \begin{tabular}{
            p{\dimexpr\linewidth-1cm-4\tabcolsep\relax}
            p{1cm}
        }
            \toprule
            \raggedright \textbf{Prior Regularization}
            & \centering \textbf{mAcc.} \tabularnewline
            \midrule
            None
            & \centering 83.7 \tabularnewline
            Uniform $(w_b=1)$
            & \centering 81.8 \tabularnewline
            Shuffled
            & \centering 76.7 \tabularnewline
            Reversed
            & \centering 80.7 \tabularnewline
            \rowcolor{blue!10}
            Slope-guided
            & \centering \textbf{85.1} \tabularnewline
            \bottomrule
        \end{tabular}
    \end{minipage}
    \hfill
    \begin{minipage}[t]{0.43\linewidth}
        \vspace{0pt}
        \centering
        \caption{Sensitivity to $\lambda_{\mathrm{prior}}$
        on AIGIBench~\cite{li2026artificial} .}
        \label{tab:lambda_prior_ablation}
        \small
        \setlength{\tabcolsep}{3pt}
        \renewcommand{\arraystretch}{1}
        \begin{tabular}{
            p{\dimexpr\linewidth-1cm-4\tabcolsep\relax}
            p{1cm}
        }
            \toprule
            \centering \textbf{$\lambda_{\mathrm{prior}}$}
            & \centering \textbf{mAcc.} \tabularnewline
            \midrule
            \centering 0.05
            & \centering 78.8 \tabularnewline
            \centering 0.075
            & \centering 82.5 \tabularnewline
            \rowcolor{blue!10}
            \centering 0.10
            & \centering \textbf{85.1} \tabularnewline
            \centering 0.125
            & \centering 81.0 \tabularnewline
            \centering 0.15
            & \centering 82.8 \tabularnewline
            \bottomrule
        \end{tabular}
    \end{minipage}
\end{table}





\subsubsection{Effect of Slope-Guided Weighting}
Table~\ref{tab:prior_ablation} shows that slope-guided weighting
improves mAcc.\ from 83.7\% to 85.1\%, while uniform weighting
reduces it to 81.8\%.
Shuffled and Reversed preserve the weight distribution and
unit batch mean but disrupt or invert the weight--slope
correspondence, reducing mAcc.\ to 76.7\% and 80.7\%, respectively.
These results support stronger prior preservation for samples
with larger slope deviations and highlight the importance
of weight--sample correspondence.

\subsubsection{Sensitivity to the Prior-Regularization Weight}
We select $\lambda_{\mathrm{prior}}$ by validation mAcc.\ on
AIGIBench~\cite{li2026artificial} and report test results in
Table~\ref{tab:lambda_prior_ablation}.
Among the evaluated weights, $\lambda_{\mathrm{prior}}=0.10$
achieves the highest mAcc.\ of 85.1\%, while both smaller
and larger values reduce accuracy.

\subsection{Robust Analysis}

Following \cite{frank2020leveraging}, we evaluate robustness under blur, cropping, JPEG compression, Gaussian noise, and their combination, each applied with a 50\% probability. As shown in Tables~\ref{tab:robustness1} and~\ref{tab:robustness2}, our method consistently achieves the best mAcc. on both benchmarks. Under combined perturbations, it reaches 78.2\%/83.0\% mAcc./AP on AIGIBench and 86.7\%/93.5\% on AIGCDetectBench, outperforming the strongest competitors by 3.7/3.5 and 3.4/2.9 points, respectively. These results demonstrate that SlopeGuard maintains state-of-the-art mean accuracy across all evaluated perturbation settings.

\section{Conclusion}
In this paper, we presented \textbf{SlopeGuard}, a Radial Spectral Slope guided framework for generalizable AI generated image detection. Our analysis revealed generator dependent distributional shifts between real and generated images, motivating Radial Spectral Slope as complementary spectral evidence rather than a universal decision threshold. SlopeGuard couples slope-aware data processing with
slope-guided feature learning to integrate spectral decay
cues from reconstructed views with visual representations
and adapt prior preservation according to spectral atypicality. Extensive experiments across multiple benchmarks demonstrate state-of-the-art detection performance and  robustness.

\section*{LLM Usage Statement}
GPT-6.1 Sol was used to assist with language editing, grammar correction, manuscript organization, and the implementation and debugging of selected code components. GPT-6.1 Sol was also used for supplementary checks of mathematical expressions and their internal consistency. The authors independently developed the mathematical formulations, conducted the experimental analyzes, and determined the scientific interpretations. All AI-assisted text and code were carefully reviewed and validated by the authors, with numerical results checked against the underlying experimental outputs. The authors take full responsibility for the study design, implementation, reported results, scientific interpretations, and final manuscript.
{
    \small
    \bibliographystyle{ieeenat_fullname}
    \bibliography{main}

@article{kingma2014adam,
  title={Adam: A method for stochastic optimization},
  author={Kingma, Diederik P and Ba, Jimmy},
  journal={arXiv preprint arXiv:1412.6980},
  year={2014}
}

@inproceedings{tan2025c2p,
  title={C2p-clip: Injecting category common prompt in clip to enhance generalization in deepfake detection},
  author={Tan, Chuangchuang and Tao, Renshuai and Liu, Huan and Gu, Guanghua and Wu, Baoyuan and Zhao, Yao and Wei, Yunchao},
  booktitle={Proceedings of the AAAI Conference on Artificial Intelligence},
  volume={39},
  number={7},
  pages={7184--7192},
  year={2025}
}

@inproceedings{ojha2023towards,
  title={Towards universal fake image detectors that generalize across generative models},
  author={Ojha, Utkarsh and Li, Yuheng and Lee, Yong Jae},
  booktitle={Proceedings of the IEEE/CVF conference on computer vision and pattern recognition},
  pages={24480--24489},
  year={2023}
}

@inproceedings{yan2025sanity,
  title={A sanity check for ai-generated image detection},
  author={Yan, Shilin and Li, Ouxiang and Cai, Jiayin and Hao, Yanbin and Jiang, Xiaolong and Hu, Yao and Xie, Weidi},
  booktitle={International Conference on Learning Representations},
  volume={2025},
  pages={70702--70720},
  year={2025}
}

@article{li2026artificial,
  title={Is artificial intelligence generated image detection a solved problem?},
  author={Li, Ziqiang and Yan, Jiazhen and He, Ziwen and Zeng, Kai and Jiang, Weiwei and Xiong, Lizhi and Fu, Zhangjie},
  journal={Advances in Neural Information Processing Systems},
  volume={38},
  year={2026}
}

@article{karras2017progressive,
  title={Progressive growing of gans for improved quality, stability, and variation},
  author={Karras, Tero and Aila, Timo and Laine, Samuli and Lehtinen, Jaakko},
  journal={arXiv preprint arXiv:1710.10196},
  year={2017}
}

@inproceedings{rombach2022high,
  title={High-resolution image synthesis with latent diffusion models},
  author={Rombach, Robin and Blattmann, Andreas and Lorenz, Dominik and Esser, Patrick and Ommer, Bj{\"o}rn},
  booktitle={Proceedings of the IEEE/CVF conference on computer vision and pattern recognition},
  pages={10684--10695},
  year={2022}
}

@article{zhong2023patchcraft,
  title={Patchcraft: Exploring texture patch for efficient ai-generated image detection},
  author={Zhong, Nan and Xu, Yiran and Li, Sheng and Qian, Zhenxing and Zhang, Xinpeng},
  journal={arXiv preprint arXiv:2311.12397},
  year={2023}
}

@article{zhu2023genimage,
  title={Genimage: A million-scale benchmark for detecting ai-generated image},
  author={Zhu, Mingjian and Chen, Hanting and Yan, Qiangyu and Huang, Xudong and Lin, Guanyu and Li, Wei and Tu, Zhijun and Hu, Hailin and Hu, Jie and Wang, Yunhe},
  journal={Advances in neural information processing systems},
  volume={36},
  pages={77771--77782},
  year={2023}
}

@inproceedings{li2026towards,
  title={Towards generalizable ai-generated image detection via image-adaptive prompt learning},
  author={Li, Yiheng and Tan, Zichang and Xu, Guoqing and Lei, Zhen and Zhou, Xu and Yang, Yang},
  booktitle={Proceedings of the IEEE/CVF Conference on Computer Vision and Pattern Recognition},
  pages={21262--21272},
  year={2026}
}

@article{dhariwal2021diffusion,
  title={Diffusion models beat gans on image synthesis},
  author={Dhariwal, Prafulla and Nichol, Alexander},
  journal={Advances in neural information processing systems},
  volume={34},
  pages={8780--8794},
  year={2021}
}

@inproceedings{wang2020cnn,
  title={CNN-generated images are surprisingly easy to spot... for now},
  author={Wang, Sheng-Yu and Wang, Oliver and Zhang, Richard and Owens, Andrew and Efros, Alexei A},
  booktitle={Proceedings of the IEEE/CVF conference on computer vision and pattern recognition},
  pages={8695--8704},
  year={2020}
}

@inproceedings{tan2024rethinking,
  title={Rethinking the up-sampling operations in cnn-based generative network for generalizable deepfake detection},
  author={Tan, Chuangchuang and Zhao, Yao and Wei, Shikui and Gu, Guanghua and Liu, Ping and Wei, Yunchao},
  booktitle={Proceedings of the IEEE/CVF conference on computer vision and pattern recognition},
  pages={28130--28139},
  year={2024}
}

@inproceedings{tan2024frequency,
  title={Frequency-aware deepfake detection: Improving generalizability through frequency space domain learning},
  author={Tan, Chuangchuang and Zhao, Yao and Wei, Shikui and Gu, Guanghua and Liu, Ping and Wei, Yunchao},
  booktitle={Proceedings of the AAAI Conference on Artificial Intelligence},
  volume={38},
  number={5},
  pages={5052--5060},
  year={2024}
}

@inproceedings{li2025improving,
  title={Improving synthetic image detection towards generalization: An image transformation perspective},
  author={Li, Ouxiang and Cai, Jiayin and Hao, Yanbin and Jiang, Xiaolong and Hu, Yao and Feng, Fuli},
  booktitle={Proceedings of the 31st ACM SIGKDD Conference on Knowledge Discovery and Data Mining V. 1},
  pages={2405--2414},
  year={2025}
}

@article{yan2026dual,
  title={Dual frequency branch framework with reconstructed sliding windows attention for ai-generated image detection},
  author={Yan, Jiazhen and Li, Ziqiang and Wang, Fan and He, Ziwen and Fu, Zhangjie},
  journal={IEEE Transactions on Information Forensics and Security},
  year={2026},
  publisher={IEEE}
}

@inproceedings{yang2026layer,
  title={Layer Consistency Matters: Elegant Latent Transition Discrepancy for Generalizable Synthetic Image Detection},
  author={Yang, Yawen and Li, Feng and Kong, Shuqi and Diao, Yunfeng and Gao, Xinjian and Shi, Zenglin and Wang, Meng},
  booktitle={Proceedings of the IEEE/CVF Conference on Computer Vision and Pattern Recognition},
  pages={38111--38121},
  year={2026}
}

@inproceedings{radford2021learning,
  title={Learning transferable visual models from natural language supervision},
  author={Radford, Alec and Kim, Jong Wook and Hallacy, Chris and Ramesh, Aditya and Goh, Gabriel and Agarwal, Sandhini and Sastry, Girish and Askell, Amanda and Mishkin, Pamela and Clark, Jack and others},
  booktitle={International conference on machine learning},
  pages={8748--8763},
  year={2021},
  organization={PmLR}
}

@inproceedings{yang2025d,
  title={D\^{} 3: scaling up deepfake detection by learning from discrepancy},
  author={Yang, Yongqi and Qian, Zhihao and Zhu, Ye and Russakovsky, Olga and Wu, Yu},
  booktitle={Proceedings of the Computer Vision and Pattern Recognition Conference},
  pages={23850--23859},
  year={2025}
}

@inproceedings{frank2020leveraging,
  title={Leveraging frequency analysis for deep fake image recognition},
  author={Frank, Joel and Eisenhofer, Thorsten and Sch{\"o}nherr, Lea and Fischer, Asja and Kolossa, Dorothea and Holz, Thorsten},
  booktitle={International conference on machine learning},
  pages={3247--3258},
  year={2020},
  organization={PMLR}
}

@inproceedings{zhou2025aigi,
  title={Aigi-holmes: Towards explainable and generalizable ai-generated image detection via multimodal large language models},
  author={Zhou, Ziyin and Luo, Yunpeng and Wu, Yuanchen and Sun, Ke and Ji, Jiayi and Yan, Ke and Ding, Shouhong and Sun, Xiaoshuai and Wu, Yunsheng and Ji, Rongrong},
  booktitle={Proceedings of the IEEE/CVF International Conference on Computer Vision},
  pages={18746--18758},
  year={2025}
}

@inproceedings{corvi2023intriguing,
  title={Intriguing properties of synthetic images: from generative adversarial networks to diffusion models},
  author={Corvi, Riccardo and Cozzolino, Davide and Poggi, Giovanni and Nagano, Koki and Verdoliva, Luisa},
  booktitle={Proceedings of the IEEE/CVF conference on computer vision and pattern recognition},
  pages={973--982},
  year={2023}
}

@inproceedings{zhang2025towards,
  title={Towards universal ai-generated image detection by variational information bottleneck network},
  author={Zhang, Haifeng and He, Qinghui and Bi, Xiuli and Li, Weisheng and Liu, Bo and Xiao, Bin},
  booktitle={Proceedings of the Computer Vision and Pattern Recognition Conference},
  pages={23828--23837},
  year={2025}
}

@article{li2026spectral,
  title={Spectral Tail Auxiliary Learning for AI-Generated Image Detection},
  author={Li, Xingyi and Zhang, Jiahui and Li, Yiheng and Cao, Yun and Wang, Wenhao},
  journal={arXiv preprint arXiv:2605.22751},
  year={2026}
}

@inproceedings{tan2023learning,
  title={Learning on gradients: Generalized artifacts representation for gan-generated images detection},
  author={Tan, Chuangchuang and Zhao, Yao and Wei, Shikui and Gu, Guanghua and Wei, Yunchao},
  booktitle={Proceedings of the IEEE/CVF Conference on Computer Vision and Pattern Recognition},
  pages={12105--12114},
  year={2023}
}

@inproceedings{durall2020watch,
  title={Watch your up-convolution: Cnn based generative deep neural networks are failing to reproduce spectral distributions},
  author={Durall, Ricard and Keuper, Margret and Keuper, Janis},
  booktitle={Proceedings of the IEEE/CVF conference on computer vision and pattern recognition},
  pages={7890--7899},
  year={2020}
}

@article{li2026reduce,
  title={Reduce the Artifacts Bias for More Generalizable AI-Generated Image Detection},
  author={Li, Yiheng and Yang, Yang and Tan, Zichang and Li, Gao and Lei, Zhen and Wang, Wenhao},
  journal={arXiv preprint arXiv:2605.14486},
  year={2026}
}

@inproceedings{karageorgiou2025any,
  title={Any-resolution ai-generated image detection by spectral learning},
  author={Karageorgiou, Dimitrios and Papadopoulos, Symeon and Kompatsiaris, Ioannis and Gavves, Efstratios},
  booktitle={Proceedings of the Computer Vision and Pattern Recognition Conference},
  pages={18706--18717},
  year={2025}
}

@article{tan2026aggregating,
  title={Aggregating Diverse Cue Experts for AI-Generated Image Detection},
  author={Tan, Lei and Li, Shuwei and Kankanhalli, Mohan and Tan, Robby T},
  journal={arXiv preprint arXiv:2601.08790},
  year={2026}
}

@article{goodfellow2014generative,
  title={Generative adversarial nets},
  author={Goodfellow, Ian J and Pouget-Abadie, Jean and Mirza, Mehdi and Xu, Bing and Warde-Farley, David and Ozair, Sherjil and Courville, Aaron and Bengio, Yoshua},
  journal={Advances in neural information processing systems},
  volume={27},
  year={2014}
}

@inproceedings{liu2026tsad,
  title={TSAD: Trace-Semantic Adaptive Disentanglement for Detecting and Grounding Multi-Modal Media Manipulation},
  author={Liu, Wenzheng and Fu, Cheng and Peng, Hujin and Wu, Junlong and Chen, Xianhong and Huang, Lan and Xu, Jing and Tian, Yixi and Liang, Menghan and Wang, Xiaofeng and others},
  booktitle={Proceedings of the 2026 International Conference on Multimedia Retrieval},
  pages={2077--2086},
  year={2026}
}

@article{yuan2026mlep,
  title={Mlep: Multi-granularity local entropy patterns for generalized ai-generated image detection},
  author={Yuan, Lin and Li, Xiaowan and Zhang, Yan and Zhang, Jiawei and Li, Hongbo and Gao, Xinbo},
  journal={Advances in Neural Information Processing Systems},
  volume={38},
  pages={68981--69000},
  year={2026}
}

@article{liang2026ferretnet,
  title={Ferretnet: Efficient synthetic image detection via local pixel dependencies},
  author={Liang, Shuqiao and Liu, Jian and Renzhang, Chen and Guan, Quanlong},
  journal={Advances in Neural Information Processing Systems},
  volume={38},
  pages={42109--42133},
  year={2026}
}

@article{yan2025dgs,
  title={DGS-Net: Distillation-Guided Gradient Surgery for CLIP Fine-Tuning in AI-Generated Image Detection},
  author={Yan, Jiazhen and Li, Ziqiang and Wang, Fan and Wang, Boyu and He, Ziwen and Fu, Zhangjie},
  journal={arXiv preprint arXiv:2511.13108},
  year={2025}
}

@inproceedings{liu2020global,
  title={Global texture enhancement for fake face detection in the wild},
  author={Liu, Zhengzhe and Qi, Xiaojuan and Torr, Philip HS},
  booktitle={Proceedings of the IEEE/CVF conference on computer vision and pattern recognition},
  pages={8060--8069},
  year={2020}
}

@inproceedings{zhou2026beyond,
  title={Beyond Semantic Features: Pixel-level Mapping for Generalized AI-Generated Image Detection},
  author={Zhou, Chenming and Wang, Jiaan and Li, Yu and Li, Lei and Cao, Juan and Tang, Sheng},
  booktitle={Proceedings of the AAAI Conference on Artificial Intelligence},
  volume={40},
  number={42},
  pages={36101--36109},
  year={2026}
}

@inproceedings{yan2025orthogonal,
  title={Orthogonal Subspace Decomposition for Generalizable AI-Generated Image Detection},
  author={Yan, Zhiyuan and Wang, Jiangming and Jin, Peng and Zhang, Ke-Yue and Liu, Chengchun and Chen, Shen and Yao, Taiping and Ding, Shouhong and Wu, Baoyuan and Yuan, Li},
  booktitle={International Conference on Machine Learning},
  pages={70268--70288},
  year={2025},
  organization={PMLR}
}

@article{huang2024gan,
  title={The gan is dead; long live the gan! a modern gan baseline},
  author={Huang, Yiwen and Gokaslan, Aaron and Kuleshov, Volodymyr and Tompkin, James},
  journal={Advances in Neural Information Processing Systems},
  volume={37},
  pages={44177--44215},
  year={2024}
}

@inproceedings{wu2024infinite,
  title={Infinite-id: Identity-preserved personalization via id-semantics decoupling paradigm},
  author={Wu, Yi and Li, Ziqiang and Zheng, Heliang and Wang, Chaoyue and Li, Bin},
  booktitle={European Conference on Computer Vision},
  pages={279--296},
  year={2024},
  organization={Springer}
}

@article{liu2026mirror,
  title={MIRROR: Manifold Ideal Reference ReconstructOR for Generalizable AI-Generated Image Detection},
  author={Liu, Ruiqi and Cui, Manni and Qin, Ziheng and Yan, Zhiyuan and Chen, Ruoxin and Han, Yi and Li, Zhiheng and Chen, Junkai and Chen, ZhiJin and Lin, Kaiqing and others},
  journal={arXiv preprint arXiv:2602.02222},
  year={2026}
}

@inproceedings{corvi2023detection,
  title={On the detection of synthetic images generated by diffusion models},
  author={Corvi, Riccardo and Cozzolino, Davide and Zingarini, Giada and Poggi, Giovanni and Nagano, Koki and Verdoliva, Luisa},
  booktitle={ICASSP 2023-2023 IEEE International Conference on Acoustics, Speech and Signal Processing (ICASSP)},
  pages={1--5},
  year={2023},
  organization={IEEE}
}

@article{wolter2022wavelet,
  title={Wavelet-packets for deepfake image analysis and detection},
  author={Wolter, Moritz and Blanke, Felix and Heese, Raoul and Garcke, Jochen},
  journal={Machine Learning},
  volume={111},
  number={11},
  pages={4295--4327},
  year={2022},
  publisher={Springer}
}

@inproceedings{xu2024ufogen,
  title={Ufogen: You forward once large scale text-to-image generation via diffusion gans},
  author={Xu, Yanwu and Zhao, Yang and Xiao, Zhisheng and Hou, Tingbo},
  booktitle={Proceedings of the IEEE/CVF Conference on Computer Vision and Pattern Recognition},
  pages={8196--8206},
  year={2024}
}

@inproceedings{jia2025chatgen,
  title={Chatgen: Automatic text-to-image generation from freestyle chatting},
  author={Jia, Chengyou and Xia, Changliang and Dang, Zhuohang and Wu, Weijia and Qian, Hangwei and Luo, Minnan},
  booktitle={Proceedings of the IEEE/CVF Conference on Computer Vision and Pattern Recognition},
  pages={13284--13293},
  year={2025}
}

@article{zheng2024breaking,
  title={Breaking semantic artifacts for generalized ai-generated image detection},
  author={Zheng, Chende and Lin, Chenhao and Zhao, Zhengyu and Wang, Hang and Guo, Xu and Liu, Shuai and Shen, Chao},
  journal={Advances in Neural Information Processing Systems},
  volume={37},
  pages={59570--59596},
  year={2024}
}

@article{yan2025ns,
  title={NS-Net: Decoupling CLIP Semantic Information through NULL-Space for Generalizable AI-Generated Image Detection},
  author={Yan, Jiazhen and Wang, Fan and Jiang, Weiwei and Li, Ziqiang and Fu, Zhangjie},
  journal={arXiv preprint arXiv:2508.01248},
  year={2025}
}

@inproceedings{wang2025lota,
  title={Lota: Bit-planes guided ai-generated image detection},
  author={Wang, Hongsong and Cheng, Renxi and Zhang, Yang and Han, Chaolei and Gui, Jie},
  booktitle={2025 IEEE/CVF International Conference on Computer Vision (ICCV)},
  pages={17246--17255},
  year={2025},
  organization={IEEE}
}

@inproceedings{chu2025fire,
  title={Fire: Robust detection of diffusion-generated images via frequency-guided reconstruction error},
  author={Chu, Beilin and Xu, Xuan and Wang, Xin and Zhang, Yufei and You, Weike and Zhou, Linna},
  booktitle={2025 IEEE/CVF Conference on Computer Vision and Pattern Recognition (CVPR)},
  pages={12830--12839},
  year={2025},
  organization={IEEE}
}

@inproceedings{yang2026all,
  title={All patches matter, more patches better: Enhance ai-generated image detection via panoptic patch learning},
  author={Yang, Zheng and Chen, Ruoxin and Yan, Zhiyuan and Zhang, Ke-Yue and Fu, Xinghe and Wu, Shuang and Shu, Xiujun and Yao, Taiping and Ding, Shouhong and Qin, Zequn and others},
  booktitle={International Conference on Learning Representations},
  volume={2026},
  pages={121742--121761},
  year={2026}
}

@inproceedings{mu2026no,
  title={No pixel left behind: A detail-preserving architecture for robust high-resolution AI-generated image detection},
  author={Mu, Lianrui and Hu, Haoji and Zou, Xingze and Bai, Jianhong and Hu, Jiaqi},
  booktitle={International Conference on Learning Representations},
  volume={2026},
  pages={80332--80355},
  year={2026}
}

@inproceedings{liu2024forgery,
  title={Forgery-aware adaptive transformer for generalizable synthetic image detection},
  author={Liu, Huan and Tan, Zichang and Tan, Chuangchuang and Wei, Yunchao and Wang, Jingdong and Zhao, Yao},
  booktitle={2024 IEEE/CVF Conference on Computer Vision and Pattern Recognition (CVPR)},
  pages={10770--10780},
  year={2024},
  organization={IEEE}
}

@inproceedings{shi2025mirage,
  title={MiraGe: Multimodal discriminative representation learning for generalizable AI-generated image detection},
  author={Shi, Kuo and Lu, Jie and Ye, Shanshan and Zhang, Guangquan and Fang, Zhen},
  booktitle={Proceedings of the 33rd ACM International Conference on Multimedia},
  pages={353--361},
  year={2025}
}

@inproceedings{zhang2025d,
  title={D 3 QE: Learning Discrete Distribution Discrepancy-Aware Quantization Error for Autoregressive-Generated Image Detection},
  author={Zhang, Yanran and Yu, Bingyao and Zheng, Yu and Zheng, Wenzhao and Duan, Yueqi and Chen, Lei and Zhou, Jie and Lu, Jiwen},
  booktitle={2025 IEEE/CVF International Conference on Computer Vision (ICCV)},
  pages={16292--16301},
  year={2025},
  organization={IEEE}
}

@article{cavia2024real,
  title={Real-time deepfake detection in the real-world},
  author={Cavia, Bar and Horwitz, Eliahu and Reiss, Tal and Hoshen, Yedid},
  journal={arXiv preprint arXiv:2406.09398},
  year={2024}
}

@inproceedings{he2026diversity,
  title={Diversity over uniformity: Rethinking representation in generated image detection},
  author={He, Qinghui and Zhang, Haifeng and Qin, Qiao and Liu, Bo and Bi, Xiuli and Xiao, Bin},
  booktitle={Proceedings of the IEEE/CVF Conference on Computer Vision and Pattern Recognition},
  pages={40407--40417},
  year={2026}
}
}

\clearpage
\setcounter{page}{1}
\maketitlesupplementary

\section{Additional Details and Analysis}

\subsection{Perturbation Settings}

Following the robustness evaluation protocol of Frank et al.~\cite{frank2020leveraging}, we consider four common image perturbations: Gaussian blur, random cropping, JPEG compression, and additive Gaussian noise. For each individual perturbation, we iterate over the entire test set and apply the corresponding corruption to each image with a probability of $50\%$. As a result, approximately half of the test images remain unchanged, while the others are perturbed, yielding a mixed evaluation setting that better reflects practical post-processing conditions.

For the combined setting, the four perturbations are considered sequentially in the order of Gaussian blur, random cropping, JPEG compression, and Gaussian noise. Each perturbation is independently applied with a probability of $50\%$, such that an image may undergo none, one, or multiple types of corruption. The detailed configurations are given below:

\begin{itemize}
    \item \textbf{Gaussian blur.}
    Gaussian filtering is applied with a kernel size randomly selected from
    $\{3,5,7,9\}$.

    \item \textbf{Random cropping.}
    The image is randomly cropped along both spatial dimensions, with the cropping ratio sampled from
    $\mathcal{U}(5,20)$. The cropped image is then resized to its original resolution.

    \item \textbf{JPEG compression.}
    JPEG compression is applied with the quality factor sampled from
    $\mathcal{U}(10,75)$, where a lower value indicates stronger compression.

    \item \textbf{Gaussian noise.}
    Independent Gaussian noise is added to the image, with its variance sampled from
    $\mathcal{U}(5.0,20.0)$.

    \item \textbf{Combined.}
    Each of the four perturbations is independently applied to the image, with a probability of 50\%.
\end{itemize}

Results are computed over the full test set, including
images left unchanged by the random perturbation process.
All perturbations are applied only at test time.
Each detector uses its corresponding pretrained checkpoint
without additional training, adaptation, or parameter tuning. 

\subsection{Details of Weighting Baselines}
\label{sec:supp_weighting}

All variants share the same training configuration and
prior-regularization coefficient of 0.10, differing only
in sample weight assignment.

\textbf{Uniform.}
All samples receive a weight of one, yielding equal
prior regularization across the mini-batch.

\textbf{Shuffled.}
At each training step, the slope-guided weights are randomly
permuted within the mini-batch. This preserves their values
while breaking their correspondence with spectral atypicality.

\textbf{Reversed.}
The original weights are reassigned in reverse order of
spectral atypicality, so that samples with larger deviations
receive smaller weights.

Shuffled and Reversed preserve the original weight
distribution and unit batch mean. Both reuse the same
calibration function and introduce no additional
hyperparameters.

\subsection{Details of slope sources.}
We train two variants separately, estimating slopes from either original images or reconstructed views. During training, each variant uses its corresponding slopes for both feature concatenation in the classifier and SGPR weight computation. During evaluation, the same slope source is retained for classification. Both variants use reconstructed views as CLIP inputs and share the same Patch Selection, Patch Shuffle, and all other training and evaluation settings, differing only in the slope source.
We compare slopes computed from original images and
reconstructed views while keeping the CLIP input unchanged.

\definecolor{slopehighlight}{RGB}{239,237,255}

\definecolor{oursbg}{RGB}{230,230,255}


\subsection{Effect of Slope Statistics}
\label{sec:slope_statistics}

We compare mini-batch slope statistics with fixed statistics
estimated from the training set for SGPR weighting.
Both variants are independently trained under identical settings,
differing only in the mean and standard deviation used to
compute spectral atypicality.
As shown in Table~\ref{tab:slope_statistics}, batch statistics
improve mAcc. by 2.6 percentage points and FAcc. by 6.2 points,
with a 0.9-point decrease in RAcc.
These results support adapting slope statistics to the current
mini-batch for more effective prior regularization.

\subsection{Effect of Prior-Regularization Strength}
\label{sec:prior_sensitivity}

We evaluate uniform and slope-guided prior regularization
over the same coefficient grid under identical training
settings. As shown in Table~\ref{tab:prior_sensitivity},
both strategies achieve their highest mAcc.\ at
$\lambda_{\mathrm{prior}}=0.10$, where slope-guided
weighting reaches 85.1\%, outperforming uniform weighting
by 3.3 percentage points. Slope-guided weighting also
performs better at four of the five tested coefficients,
although its performance varies with regularization
strength.  

\begin{table*}[t]
    \centering
    \caption{Comparison with state-of-the-art methods on
ARForensics~\cite{zhang2025d} in terms of accuracy and
average precision (\%).
The top two entries are highlighted in \textbf{bold} and
\underline{underlined}, respectively.
Results for competing methods are taken from those reported
in the NS-Net paper~\cite{yan2025ns}. All models follow the same AIGIBench  training protocol~\cite{li2026artificial}.}
    \label{tab:arforensics}
    \scriptsize
    \setlength{\tabcolsep}{3pt}
    \renewcommand{\arraystretch}{1.0}
    \resizebox{\textwidth}{!}{
    \begin{tabular}{l *{7}{cc} | cc}
        \toprule
        \multirow{2}{*}{Method}
        & \multicolumn{2}{c}{LlamaGen}
        & \multicolumn{2}{c}{Janus-Pro}
        & \multicolumn{2}{c}{VAR}
        & \multicolumn{2}{c}{Infinity}
        & \multicolumn{2}{c}{Open-MAGVIT2}
        & \multicolumn{2}{c}{Switti}
        & \multicolumn{2}{c}{RAR}
        & \multicolumn{2}{c}{Mean} \\
        \cmidrule(lr){2-3}
        \cmidrule(lr){4-5}
        \cmidrule(lr){6-7}
        \cmidrule(lr){8-9}
        \cmidrule(lr){10-11}
        \cmidrule(lr){12-13}
        \cmidrule(lr){14-15}
        \cmidrule(lr){16-17}
        & Acc. & AP & Acc. & AP & Acc. & AP & Acc. & AP
        & Acc. & AP & Acc. & AP & Acc. & AP & mAcc. & mAP \\
        \midrule
        CNN-Spot~\cite{wang2020cnn}
        & 50.9 & 72.4 & 63.3 & 91.6 & 50.1 & 61.5 & 63.4 & 94.5
        & 50.2 & 58.2 & 54.3 & 81.9 & 50.6 & 68.7 & 54.7 & 75.5 \\
        UniFD~\cite{ojha2023towards}
        & 71.2 & 85.0 & 80.7 & 91.3 & 61.1 & 74.9 & 91.8 & 97.0
        & 59.3 & 72.7 & 91.2 & 96.9 & 83.2 & 93.2 & 76.9 & 87.3 \\
        FreqNet~\cite{tan2024frequency}
        & 90.3 & 96.9 & 95.6 & 99.8 & 78.3 & 89.2 & 95.5 & 99.7
        & 76.4 & 88.3 & 95.3 & 99.8 & 80.8 & 91.7 & 87.5 & 95.1 \\
        NPR~\cite{tan2024rethinking}
        & 83.5 & 97.5 & 97.0 & 99.8 & 81.8 & 98.6 & 89.7 & 99.5
        & 55.7 & 82.0 & 99.4 & 100.0 & 67.7 & 93.6 & 82.1 & 95.9 \\
        LaDeDa~\cite{cavia2024real}
        & 89.4 & 99.6 & 99.6 & 100.0 & 80.5 & 98.7 & 99.8 & 100.0
        & 78.0 & 99.0 & 99.8 & 100.0 & 87.1 & 99.6 & 90.6 & 99.5 \\
        AIDE~\cite{yan2025sanity}
        & 91.0 & 97.8 & 96.3 & 99.4 & 92.8 & 98.4 & 95.3 & 98.8
        & \underline{93.4} & 98.6 & 94.3 & 99.0
        & 92.5 & 98.4 & 93.7 & 98.6 \\
        DFFreq~\cite{yan2026dual}
        & 96.2 & 99.9 & 99.2 & 100.0 & 91.1 & 99.8 & 96.9 & 99.9
        & 78.4 & 99.0 & 99.5 & 100.0 & 96.6 & 100.0 & 94.0 & 99.8 \\
        SAFE~\cite{li2025improving}
        & 96.4 & \underline{100.0}
        & 99.8 & \underline{100.0}
        & \underline{99.1} & \underline{100.0}
        & \underline{99.8} & \underline{100.0}
        & 91.4 & \textbf{99.9}
        & \underline{99.9} & 100.0
        & 96.7 & 100.0
        & 97.6 & \textbf{100.0} \\
        VIB-Net~\cite{zhang2025towards}
        & 86.2 & 96.1 & 98.4 & 99.7 & 67.0 & 85.9 & 98.9 & 99.9
        & 75.1 & 90.9 & 95.3 & 99.3 & 93.2 & 98.4 & 87.7 & 95.8 \\
        Effort~\cite{yan2025orthogonal}
        & 97.5 & 100.0
        & \underline{100.0} & 100.0
        & 92.5 & 99.8 & 99.6 & 100.0
        & 81.8 & 97.8 & 98.4 & 100.0
        & \underline{99.8} & 100.0 & 95.7 & 99.7 \\
        LTD~\cite{yang2026layer}
        & 95.6 & 99.9 & 99.8 & 100.0 & 85.2 & 99.5 & 99.7 & 100.0
        & 80.6 & 99.1 & 98.2 & 99.9 & 99.6 & 100.0 & 94.1 & 99.8 \\
        DoU~\cite{he2026diversity}
        & 98.3 & \textbf{100.0}
        & \textbf{100.0} & 100.0
        & 94.4 & 99.9 & 99.8 & 100.0
        & 84.0 & \underline{99.7} & 99.0 & 100.0
        & 98.9 & 100.0 & 96.3 & 99.9 \\
        NS-Net~\cite{yan2025ns}
        & \underline{99.2} & 99.9
        & 99.9 & \textbf{100.0}
        & 98.1 & 99.8
        & \textbf{100.0} & \textbf{100.0}
        & 91.2 & 98.0
        & \textbf{99.9} & \underline{100.0}
        & \textbf{99.9} & \underline{100.0}
        & \underline{98.3} & 99.7 \\
        \midrule
        \rowcolor{blue!10}
        Ours
        & \textbf{99.6} & 99.9
        & 99.6 & 99.9
        & \textbf{99.7} & \textbf{100.0}
        & 99.7 & 99.9
        & \textbf{95.8} & 99.5
        & 99.6 & \textbf{100.0}
        & 99.6 & \textbf{100.0}
        & \textbf{99.1} & \underline{99.9} \\
        \bottomrule
    \end{tabular}
    }
\end{table*}

\begin{table}[t]
    \centering
    \caption{Effect of slope statistics on AIGIBench (\%)~\cite{li2026artificial}.
    Fixed statistics are estimated from the training set and
    remain unchanged during training.}
    \label{tab:slope_statistics}
    \small
    \renewcommand{\arraystretch}{1.1}
    \begin{tabularx}{\columnwidth}{
        >{\raggedright\arraybackslash}X c c c
    }
        \toprule
        Statistics & RAcc. & FAcc. & mAcc. \\
        \midrule
        Fixed training & 90.3 & 74.6 & 82.5 \\
        \rowcolor{blue!10}
        Batch & 89.4 & 80.8 & \textbf{85.1} \\
        \bottomrule
     \end{tabularx}
\end{table}

\begin{table}[t]
    \centering
    \caption{Effect of prior-regularization strength on
    AIGIBench (\%)~\cite{li2026artificial}. Both weighting strategies share the same training settings. Results are reported in mAcc.}
    \label{tab:prior_sensitivity}
    \small
    \setlength{\tabcolsep}{3pt}
    \renewcommand{\arraystretch}{1.12}
    \begin{tabular*}{\columnwidth}{
        @{\extracolsep{\fill}}lccccc@{}
    }
        \toprule
        $\lambda_{\mathrm{prior}}$
        & 0.05 & 0.075 & 0.10 & 0.125 & 0.15 \\
        \midrule
        Uniform
        & 80.9 & 80.6 & 81.8 & 79.4 & 79.4 \\
        
        Slope-guided
        & 78.8 & 82.5 & \cellcolor{blue!10}\textbf{85.1} & 81.0 & 82.8 \\
        \bottomrule
    \end{tabular*}
\end{table}

\begin{table}[t]
    \centering
    \caption{Comparison of scalar spectral descriptors on
    AIGIBench~\cite{li2026artificial} (\%). SADP is enabled and
    SGPR is disabled in all variants. Each descriptor is
    concatenated with the CLIP representation under identical
    training settings.}
    \label{tab:spectral_descriptors}
    \small
    \setlength{\tabcolsep}{5pt}
    \renewcommand{\arraystretch}{1.12}
    \begin{tabular}{lc}
        \toprule
        Spectral Descriptor & mAcc. \\
        \midrule
        None & 78.8 \\
        Frequency energy ratio & 80.3 \\
        Spectral entropy & 81.8 \\
        \rowcolor{blue!10}
        Radial spectral slope & \textbf{83.7} \\
        \bottomrule
    \end{tabular}
\end{table}

\begin{table}[t]
    \centering
    \caption{Ablation of Frequency Diversification  and
    Patch Selection  on AIGIBench (\%)~\cite{li2026artificial}.
    RSS and SGPR are disabled in all variants.}
    \label{tab:sadp_ablation}
    \small
    \setlength{\tabcolsep}{8pt}
    \renewcommand{\arraystretch}{1.08}
    \begin{tabular}{ccc}
        \toprule
        Frequency Diversification & Patch Selection & mAcc. \\
        \midrule
        $\times$ & $\times$ & 72.0 \\
        $\times$ & $\checkmark$ & 73.5 \\
        $\checkmark$ & $\times$ & 76.9 \\
        \rowcolor{blue!10}
        $\checkmark$ & $\checkmark$ & \textbf{78.8} \\
        \bottomrule
    \end{tabular}
\end{table}

\begin{table}[t]
    \centering
    \caption{Ablation of directional and energy consistency
    in SGPR on AIGIBench (\%)~\cite{li2026artificial}. SADP and RSS are enabled
    in all variants.}
    \label{tab:direction_energy}
    \small
    \setlength{\tabcolsep}{8pt}
    \renewcommand{\arraystretch}{1.1}
    \begin{tabular}{ccc}
        \toprule
        Direction & Energy & mAcc. \\
        \midrule
        $\times$ & $\times$ & 83.7 \\
        $\times$ & $\checkmark$ & 81.2 \\
        $\checkmark$ & $\times$ & 83.8 \\
        \rowcolor{blue!10}
        $\checkmark$ & $\checkmark$ & \textbf{85.1} \\
        \bottomrule
    \end{tabular}
\end{table}

\subsection{Comparison of Spectral Descriptors}
\label{sec:spectral_descriptors}
We compare RSS with frequency energy ratio and spectral
entropy, each concatenated as a scalar with the CLIP
representation under identical training settings.
SADP is enabled and SGPR is disabled throughout.
As shown in Table~\ref{tab:spectral_descriptors},
all three descriptors improve mAcc.\ over the baseline
without spectral features. RSS achieves the highest
mAcc.\ of 83.7\%, exceeding frequency energy ratio and
spectral entropy by 3.4 and 1.9 percentage points,
respectively. These results support RSS as an effective
source of complementary spectral information.

\subsection{Component Analysis of SADP}
\label{sec:sadp_ablation}
Table~\ref{tab:sadp_ablation} evaluates the individual and combined
effects of FD and PS with RSS and SGPR disabled.
PS alone yields a modest improvement of 1.5 points in mAcc.\
over the baseline.
FD provides a larger improvement, increasing mAcc.\ from 72.0\%
to 76.9\%.
Combining both components achieves the best mAcc.\ of 78.8\%,
representing a further gain of 1.9 points over FD alone.
These results demonstrate that frequency diversification and
patch selection complement each other, leading to more effective
slope-aware data processing.

\subsection{Component Analysis of SGPR}
\label{sec:direction_energy}
We ablate the directional and energy consistency terms in
SGPR to assess their individual contributions and
complementarity. All variants are independently trained
with identical SADP, RSS, and optimization settings.
As shown in Table~\ref{tab:direction_energy}, energy
consistency alone reduces mAcc.\ from 83.7\% to 81.2\%,
while directional consistency alone yields a marginal
improvement to 83.8\%. Combining both terms achieves
the highest mAcc.\ of 85.1\%, outperforming directional
consistency alone by 1.3 percentage points.
These results suggest that energy consistency is
beneficial when coupled with directional alignment,
highlighting their complementary roles in prior preservation.


\subsection{Evaluation on ARForensics}
\label{sec:arforensics}

We further evaluate generalization to autoregressive image
generation on ARForensics~\cite{zhang2025d}, which covers
seven generators, using baseline results reported in
NS-Net~\cite{yan2025ns}. As shown in Table~\ref{tab:arforensics},
our method achieves the highest mAcc.\ of 99.1\%, surpassing
NS-Net and SAFE by 0.8 and 1.5 percentage points, respectively,
while attaining 99.9\% mAP. It achieves the best accuracy on
three generators, improving over NS-Net by 0.4, 1.6, and
4.6 percentage points on LlamaGen, VAR, and Open-MAGVIT2,
respectively. Moreover, accuracy remains at least 99.6\%
on six generators, with 95.8\% on Open-MAGVIT2, providing
further evidence of generalization across diverse
image-generation paradigms.

\begin{table}[t]
\centering\footnotesize
\setlength{\tabcolsep}{2pt}
\renewcommand{\arraystretch}{1.12}
\caption{Reconstructed-view RSS descriptive statistics. All groups contain 1000 images. SD: sample standard deviation; IQR: interquartile range.}
\label{tab:rss-descriptive}
\begin{tabular*}{\columnwidth}{@{\extracolsep{\fill}}lrrrr@{}}
\toprule
\textbf{Group} & \textbf{Mean} & \textbf{SD} & \textbf{Median} & \textbf{IQR} \\
\midrule
\rowcolor{blue!10}
Shared real & -0.02339 & 0.00624 & -0.02292 & 0.00851 \\
ADM & -0.02512 & 0.00621 & -0.02491 & 0.00894 \\
DALL-E 2 & -0.02942 & 0.00292 & -0.02967 & 0.00352 \\
Glide & -0.03958 & 0.00368 & -0.03998 & 0.00453 \\
Midjourney & -0.02726 & 0.00794 & -0.02973 & 0.01229 \\
VQDM & -0.02305 & 0.00619 & -0.02255 & 0.00786 \\
BigGAN & -0.02507 & 0.00367 & -0.02490 & 0.00472 \\
CycleGAN & -0.03013 & 0.00307 & -0.02975 & 0.00350 \\
GauGAN & -0.02443 & 0.00315 & -0.02454 & 0.00398 \\
ProGAN & -0.02572 & 0.00317 & -0.02587 & 0.00385 \\
SDXL & -0.02602 & 0.00335 & -0.02628 & 0.00446 \\
SD 1.4 & -0.02690 & 0.00472 & -0.02714 & 0.00637 \\
SD 1.5 & -0.02722 & 0.00495 & -0.02765 & 0.00650 \\
StarGAN & -0.03490 & 0.00146 & -0.03506 & 0.00188 \\
StyleGAN & -0.02493 & 0.00403 & -0.02567 & 0.00473 \\
StyleGAN2 & -0.02462 & 0.00510 & -0.02549 & 0.00622 \\
WhichFaceIsReal & -0.02740 & 0.00213 & -0.02756 & 0.00259 \\
Wukong & -0.02487 & 0.00611 & -0.02435 & 0.00913 \\
\bottomrule
\end{tabular*}
\end{table}

\begin{table*}[t]
\centering\small
\setlength{\tabcolsep}{4pt}
\renewcommand{\arraystretch}{1}
\caption{Generator-level RSS differences on reconstructed views against one shared real reference. All groups contain 1000 images. CvM is the SciPy two-sample statistic without additional rescaling; $\Delta_{50}$ is the generated minus real median.}
\label{tab:rss-main}
\begin{tabular*}{\textwidth}{@{\extracolsep{\fill}}lrrrrrr@{}}
\toprule
\textbf{Generator} & \textbf{CvM} & \textbf{KS} & $W_1$ & $\Delta_{50}$ & $p$ & $p_{\mathrm{Holm}}$ \\
\midrule
ADM & 3.867 & 0.132 & 0.00173 & -0.00199 & 0.0001 & 0.0017 \\
DALL-E 2 & 64.564 & 0.554 & 0.00620 & -0.00674 & 0.0001 & 0.0017 \\
Glide & 152.573 & 0.890 & 0.01618 & -0.01706 & 0.0001 & 0.0017 \\
Midjourney & 19.418 & 0.338 & 0.00415 & -0.00681 & 0.0001 & 0.0017 \\
VQDM & 0.216 & 0.042 & 0.00040 & +0.00038 & 0.2441 & 0.2441 \\
BigGAN & 11.104 & 0.248 & 0.00257 & -0.00198 & 0.0001 & 0.0017 \\
CycleGAN & 74.279 & 0.628 & 0.00674 & -0.00683 & 0.0001 & 0.0017 \\
GauGAN & 10.671 & 0.231 & 0.00284 & -0.00162 & 0.0001 & 0.0017 \\
ProGAN & 19.874 & 0.337 & 0.00337 & -0.00295 & 0.0001 & 0.0017 \\
SDXL & 20.738 & 0.330 & 0.00347 & -0.00335 & 0.0001 & 0.0017 \\
SD 1.4 & 21.532 & 0.313 & 0.00363 & -0.00421 & 0.0001 & 0.0017 \\
SD 1.5 & 24.253 & 0.335 & 0.00392 & -0.00473 & 0.0001 & 0.0017 \\
StarGAN & 138.597 & 0.864 & 0.01153 & -0.01214 & 0.0001 & 0.0017 \\
StyleGAN & 11.444 & 0.245 & 0.00272 & -0.00274 & 0.0001 & 0.0017 \\
StyleGAN2 & 6.597 & 0.203 & 0.00194 & -0.00257 & 0.0001 & 0.0017 \\
WhichFaceIsReal & 46.335 & 0.504 & 0.00497 & -0.00464 & 0.0001 & 0.0017 \\
Wukong & 2.698 & 0.104 & 0.00155 & -0.00143 & 0.0001 & 0.0017 \\
\bottomrule
\end{tabular*}
\end{table*}

\begin{table*}[t]
\centering
\small
\setlength{\tabcolsep}{5pt}
\renewcommand{\arraystretch}{1}
\caption{RSS distribution comparisons across image views and reference settings.
Generators are weighted equally. Significant denotes comparisons surviving
Holm correction at the 0.05 level.}
\label{tab:rss-controls}
\begin{tabular*}{\textwidth}{@{\extracolsep{\fill}}lrrrrr@{}}
\toprule
\textbf{Reference} & $n_R/n_G$ & \textbf{Mean CvM} & \textbf{Mean} $W_1$ & \textbf{95\% CI} & \textbf{Significant} \\
\midrule
\multicolumn{6}{@{}l}{\textbf{Reconstructed}} \\
\rowcolor{blue!10}
Shared real & 1000/1000 & 36.986 & 0.00458 & [0.00433, 0.00488] & 16/17 \\
JPEG matched & 665/1000 & 57.550 & 0.00600 & [0.00560, 0.00637] & 1/1 \\
PNG matched & 335/1000 & 12.092 & 0.00331 & [0.00320, 0.00351] & 16/16 \\
Directory matched & 1000/1000 & 49.150 & 0.00470 & -- & 16/17 \\
\addlinespace[3pt]
\multicolumn{6}{@{}l}{\textbf{Original (native)}} \\
Shared real & 1000/1000 & 76.836 & 0.00909 & [0.00894, 0.00927] & 17/17 \\
JPEG matched & 665/1000 & 93.292 & 0.00431 & [0.00396, 0.00470] & 1/1 \\
PNG matched & 335/1000 & 35.937 & 0.00776 & [0.00749, 0.00810] & 15/16 \\
Directory matched & 1000/1000 & 66.486 & 0.00588 & -- & 16/17 \\
\addlinespace[3pt]
\multicolumn{6}{@{}l}{\textbf{Original ($224\times224$)}} \\
Shared real & 1000/1000 & 38.829 & 0.00474 & [0.00454, 0.00496] & 17/17 \\
JPEG matched & 665/1000 & 35.675 & 0.00366 & [0.00334, 0.00399] & 1/1 \\
PNG matched & 335/1000 & 16.434 & 0.00427 & [0.00404, 0.00456] & 15/16 \\
Directory matched & 1000/1000 & 36.475 & 0.00394 & -- & 15/17 \\
\bottomrule
\end{tabular*}
\par\smallskip
\begin{minipage}{\textwidth}
\footnotesize
CIs use 2,000 bootstrap replicates; directory-reference CIs are omitted
because real pools overlap. Format subsets require at least 50 images per
group and differ in sample size and generator coverage. CvM is sample-size dependent.
\end{minipage}
\end{table*}

\section{Analysis of Radial Spectral Slope Distributions}
\label{sec:supp-radial-slope}

We compare RSS distributions across real images and 17
generators, and examine how patch reconstruction affects
these differences.

\subsection{Slope Estimation}

We reverse the input normalization, clip RGB values to
$[0,1]$, and convert the image to grayscale using weights
$(0.299,0.587,0.114)$. Let $A_k$ be the amplitude of the
orthonormal two-dimensional real FFT,
$y_k=\log(A_k+\epsilon)$, and $r_k$ the radial frequency
in discrete frequency-index units. We compute
\begin{equation}
\alpha(I)=
\frac{N^{-1}\sum_{k\in\mathcal B}
(r_k-\bar r)(y_k-\bar y)}
{N^{-1}\sum_{k\in\mathcal B}
(r_k-\bar r)^2+\epsilon},
\label{eq:radial-slope}
\end{equation}
where $N=|\mathcal B|$, $\bar r$ and $\bar y$ denote
the corresponding means, and $\epsilon=10^{-6}$.
The estimator fits individual coefficients of the one-sided
spectrum without radial averaging. The reference band
$(0,185)$ at size 256 covers the full spectrum, including
DC, after clipping and scaling its upper bound.

The main analysis uses $224\times224$ reconstructed views
from the test pipeline. Images are divided into
$32\times32$ patches, with incomplete borders discarded.
Up to 49 patches are selected using entropy extremes,
center regions, and neighboring patches; the center and
neighbor proportions are 0.55 and 0.25.
Final Patch Shuffle is disabled at test time, and the
remaining sampling randomness is fixed per image.
Ten real images have fewer than 49 complete patches;
we retain their partially black reconstructions.
For comparison, we also compute RSS on the same original
images at native resolution and after direct bicubic
resizing to $224\times224$ with antialiasing.


\subsection{Statistical Protocol}

After excluding one corrupted file and removing exact
RGB-pixel duplicates, we sample 1000 images from each
generator and 1000 real images as a shared reference.
This common size is limited by the smallest generated
group after deduplication. Sampling is without replacement
using seed 100, with proportional allocation across
available source and semantic-directory strata.
All views use the same images, and no RSS outliers
are removed.

We measure distributional differences using Cram\'er--von
Mises (CvM), Kolmogorov--Smirnov (KS), Wasserstein-1
($W_1$), and the median shift
$\Delta_{50}=\operatorname{median}(\alpha_{\mathrm{fake}})
-\operatorname{median}(\alpha_{\mathrm{real}})$.
CvM significance is assessed with 9,999 label permutations,
using $p=(1+\#\{T^{(b)}\geq T_{\mathrm{obs}}\})/10000$,
where $T_{\mathrm{obs}}$ and $T^{(b)}$ are the observed
and permuted statistics. Holm correction is applied across
generators within each setting at the 0.05 level.
Descriptive statistics are provided in
Table~\ref{tab:rss-descriptive}.

Aggregate statistics weight generators equally.
We estimate 95\% percentile confidence intervals (CIs)
for mean $W_1$ using 2,000 bootstrap replicates.
Each replicate resamples the shared real reference once
for all comparisons and resamples each generated group
independently.

\subsection{Distributional Differences}

Table~\ref{tab:rss-main} shows significant reconstructed-view
differences for 16 of 17 generators, with mean
$W_1=0.00458$ (95\% CI: [0.00433, 0.00488]).
These generators have more negative median slopes than
the real reference. 

The degree of separation varies considerably.
Glide yields the largest KS distance of 0.890,
whereas ADM and Wukong yield only 0.132 and 0.104,
respectively, despite significant tests.
RSS thus captures distributional differences across
most evaluated generators, but provides a weaker cue
for some generators.

\subsection{Sensitivity Analysis}

\paragraph{Image reconstruction.}
As shown in Table~\ref{tab:rss-controls}, all 17 generators
differ significantly from the shared reference at native
resolution and after direct resizing.
Most differences remain detectable after reconstruction,
although their direction and magnitude can change.
For example, Midjourney, SDXL, SD~1.4, WhichFaceIsReal,
and Wukong shift from positive median differences at
native resolution to negative differences after reconstruction.
Compared with directly resized images, reconstruction
changes the paired median slope by $+0.00695$ for
StarGAN and $-0.00198$ for SD~1.5.
Its effect therefore varies across generators.
Native-resolution magnitudes also depend on image size
because RSS uses frequency-index units.

\paragraph{File format and reference choice.}
Format-matched reconstructed views retain significant
differences for all 16 generators in the PNG subset
and for WhichFaceIsReal in the JPEG subset.
These comparisons use 335 and 665 real images,
respectively, and 1000 generated images per generator.
Only combinations with at least 50 images per group
are included.

Using real references from the corresponding generator
directories, with counts matched and permutations restricted
within semantic-directory strata, again yields 16 significant
comparisons. However, GauGAN replaces VQDM as the
non-significant case (adjusted $p=0.2602$).
The observed differences therefore depend partly on
the real reference.

\end{document}